\documentclass[10pt,twocolumn,letterpaper]{article}

\usepackage{3dv}
\usepackage{multirow}
\usepackage{times}
\usepackage{epsfig}
\usepackage{graphicx}
\usepackage{amsmath}
\usepackage{amssymb}
\usepackage{xcolor}
\usepackage{array, makecell}
\usepackage{multirow}
\usepackage{float}
\usepackage{gensymb}
\usepackage{subcaption}
\usepackage{mathtools}
\usepackage{amsthm}
\usepackage{booktabs}

\restylefloat{table}
\restylefloat{figure}

\usepackage{booktabs}

\DeclareMathOperator*{\argmax}{arg\,max}
\DeclareMathOperator*{\argmin}{arg\,min}
\DeclareMathOperator*{\simsup}{\sim}

\definecolor{DarkPink}{rgb}{1.0, 0.0, 1.0}

\definecolor{DarkBlue}{rgb}{0.0, 0.0, 1.0}

\definecolor{DarkGreen}{rgb}{0.0, 0.2, 0.0}

\newcommand\blfootnote[1]{%
	\begingroup
	\renewcommand\thefootnote{}\footnote{#1}%
	\addtocounter{footnote}{-1}%
	\endgroup
}

\usepackage[pagebackref=true,breaklinks=true,letterpaper=true,colorlinks,bookmarks=false]{hyperref}
\newcommand{\icg}[2]{\includegraphics[width=#1\textwidth]{#2}}

\threedvfinalcopy 

\def\threedvPaperID{202} 

\ifthreedvfinal\fi
\begin{document}

\title{On Object Symmetries and 6D Pose Estimation from Images}

\author{Giorgia Pitteri$^{1,*}$ $\quad \quad$ Micha\"el 
Ramamonjisoa$^{1,*}$ $\quad 
\quad$ Slobodan Ilic$^{2,3}$ $\quad \quad$
Vincent Lepetit$^1$ \\
	\hspace{0em} 	$^1$Laboratoire Bordelais de Recherche Informatique, 
	Universit\'e de Bordeaux, Bordeaux, France\\
	$^2$ Technische Univers\"at M\"unchen, Germany $\quad\quad$ $^3$ Siemens 
	AG, M\"unchen, Germany \\	
	{\hspace{-0em} $^1$$\texttt{\small 
			\{first.lastname\}@u-bordeaux.fr} \quad\quad$
			$^2$$\texttt{\small Slobodan.Ilic@in.tum.de}$}		 
}

\maketitle

\begin{abstract}

Objects with symmetries are common in our daily life and in industrial contexts,
but  are often  ignored in  the  recent literature  on 6D  pose estimation  from
images.  In  this paper,  we study  in an  analytical way  the link  between the
symmetries  of  a  3D object  and  its  appearance  in  images. We  explain  why
symmetrical objects can be a challenge when training machine learning algorithms
that aim at  estimating their 6D pose  from images. We propose  an efficient and
simple solution  that relies  on the  normalization of  the pose  rotation.  Our
approach   is  general   and  can   be  used   with  any   6D  pose   estimation
algorithm. Moreover, our method is also  beneficial for objects that are 'almost
symmetrical', \emph{i.e.}  objects for which  only a detail breaks the symmetry.
We validate our  approach within a Faster-RCNN framework on  a synthetic dataset
made  with objects  from  the T-Less  dataset, which  exhibit  various types  of
symmetries, as  well as  real sequences  from T-Less.
\blfootnote{$^*$ Authors with equal participation.}

\end{abstract}


\newcommand{\bhC}{\widehat{\boldsymbol{C}}}
\newcommand{\hCi}{\widehat{C}_i}
\newcommand{\bhD}{\widehat{\boldsymbol{D}}}
\newcommand{\hDi}{\widehat{D}_i}
\newcommand{\bhN}{\widehat{\boldsymbol{N}}}
\newcommand{\bhNi}{\widehat{\boldsymbol{N}}_i}

\newcommand{\bC}{\boldsymbol{C}}
\newcommand{\bD}{\boldsymbol{D}}
\newcommand{\bI}{\boldsymbol{I}}
\newcommand{\bN}{\boldsymbol{N}}
\newcommand{\bM}{\boldsymbol{M}}
\newcommand{\bX}{\boldsymbol{X}}
\newcommand{\bm}{\boldsymbol{m}}
\newcommand{\bn}{\boldsymbol{n}}
\newcommand{\bp}{\boldsymbol{p}}
\newcommand{\bu}{{\boldsymbol{u}}}
\newcommand{\bv}{{\boldsymbol{v}}}
\newcommand{\bx}{{\boldsymbol{x}}}
\newcommand{\by}{{\boldsymbol{y}}}
\newcommand{\bz}{{\boldsymbol{z}}}

\newcommand{\calC}{\mathcal{C}}
\newcommand{\calL}{\mathcal{L}}
\newcommand{\calM}{\mathcal{M}}
\newcommand{\calR}{\mathcal{R}}
\newcommand{\calX}{\mathcal{X}}
\newcommand{\calF}{\mathcal{F}}
\newcommand{\frakR}{\mathfrak{R}}

\newcommand{\Render}{\calR}
\newcommand{\map}{\text{Map}}

\newcommand{\normals}{\text{n}}
\newcommand{\depth}{\text{d}}
\newcommand{\boundaries}{\text{c}}
\newcommand{\superv}{\text{superv}}
\newcommand{\dc}{\text{dc}}
\newcommand{\dn}{\text{dn}}
\newcommand{\BerHu}{\text{BerHu}}
\newcommand{\Huber}{\text{Huber}}
\newcommand{\AL}{\text{AL}}

\newcommand{\SE}{\text{SE}}
\newcommand{\SO}{\text{SO}}

\newcommand{\frakS}{\mathfrak{S}}

\newcommand{\IR}{\mathbb{R}}
\newcommand{\IN}{\mathbb{N}}
\newcommand{\Id}{I_3}

\newcommand{\simM}{\simsup\limits^\map}
\newcommand{\Mapto}{\xrightarrow{\map}}
\newcommand{\Mappto}{\xrightarrow{\map'}}
\newcommand{\W}{\Omega}
\newcommand{\tr}[1]{\text{Trace}(#1)}
\newcommand{\Ca}{\cos(\alpha)}
\newcommand{\Sa}{\sin(\alpha)}

\newcommand{\supeq}[1]{\stackrel{\mathclap{\normalfont\mbox{#1}}}{=}}

\section{Introduction}
\label{sec:intro}
  
3D object detection and pose estimation are of primary importance for tasks such
as robotic  manipulation, virtual and augmented  reality and they have  been the
focus of  intense research  in recent years,  mostly due to  the advent  of Deep
Learning  based approaches  and  the  possibility of  using  large datasets  for
training such methods~\cite{Hinterstoisser12,drost2010CVPR,Kehl17,Rad17,
  Tekin2018,Jafari2018,Xiang18,Peng19}.

However, one  challenge is often  ignored in recent  works. Many objects  of our
daily  life  or  from  industrial  contexts  exhibit  symmetries,  or  at  least
'quasi-symmetries'  when only  a  small detail  prevents the  object  to have  a
perfect symmetry.  These  symmetries create ambiguities when  aiming to estimate
the 6D  pose of the object  from images, however  only a few recent  papers have
considered    the   problems    raised    by   object    symmetries~\cite{Rad17,
  Sundermeyer18,Bregier:hal-01415027,Manhardt18}.   In  this   paper,  we  first
explain  why  exactly  symmetries  can  be a  problem  for  6D  pose  estimation
algorithms.   We then  provide a  simple  solution that  is general  and can  be
introduced in any 6D pose estimation algorithms.

\begin{figure}
	\begin{center}
		\includegraphics[width=0.8\linewidth, height=8em]{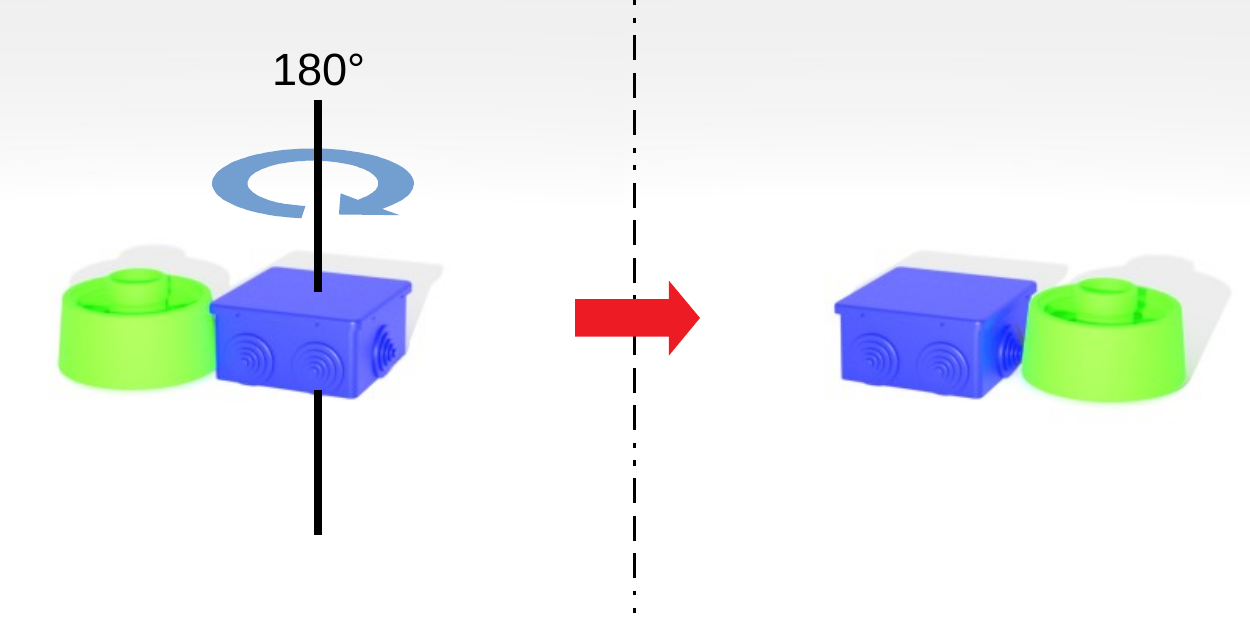}
	\caption{Two views  of the  same scene  before and  after a  rotation of
          $180\degree$ around the vertical axis  of the blue object.  Since this
          object is  symmetrical, it  has the  same appearance  but its  pose is
          different.}
	\label{fig:ambiguity_intro}
	\end{center}
\end{figure}

To better  understand the problem raised  by the symmetries of  an object, let's
first consider Fig.~\ref{fig:ambiguity_intro}.  The blue object has a rotational
symmetry around the vertical axis: If we apply a rotation of $180\degree$ around
this axis, this object has exactly the same appearance.  More generally, when an
object $O$ has some  symmetry, there exist one or more  rigid motions such that,
if we apply these rigid motions to the object pose, the appearance of the object
is preserved. Formally, we consider the set
\begin{equation}
	\begin{aligned}
		\calM = \{ &\bm \in \SE(3) \text{ such that }\\
		\quad\quad   &\forall \bp \in \SE(3), \>\>\> 
		\Render(O,\bp)=\Render(O,\bm.\bp) \} \>, 
	\end{aligned}
	\label{eq:calM}
\end{equation}
where $\Render(O,\bp)$  is the image  of Object  $O$ under pose  $\bp$ (ignoring
lighting  effects),  $\bm$ is  a  rigid  motion  related  to the  symmetry,  and
$\bm.\bp$ is the  pose after applying motion $\bm$.  $\calM(O)$  is thus the set
of rigid motions $\bm$ that preserve the visual aspect of a given object.  It is
easy to  see that  it forms a  subgroup of  $SE(3)$. \cite{Bregier:hal-01415027}
calls the elements of $\calM(O)$ \textit{proper symmetries}.

In other  words, two  images of a  symmetrical object can  be identical  but not
correspond to the same pose.  If we consider an image $\bI_1 = 
\Render(O, \bp)$ of an object  $O$ under  pose $P$  and a motion  
$\bm\in\calM(O)$, then,  the image $\bI_2$ of  object $O$ under  pose 
$\bm.\bp$ is equal  to image $\bI_1$, \emph{i.e.}  $\bI_2 =
\Render(O, \bm . \bp) = \Render(O, \bp)  = \bI_1$.  There is therefore 
no function
\begin{equation}
\calF: \bI \mapsto \bp
\label{eq:image_to_pose}
\end{equation}
that can provide the  pose $\bp$ of object $O$ given an  image $\bI$.  
Any attempt to learn  such a  function, for example  with a Deep  Network, 
would  fail.  For example, if  a network  is trained to  predict the pose  
using the  squared loss between the ground truth  poses and the predicted 
poses, it  would converge to a model predicting the average of the possible  
poses for an input image, which is of course meaningless.

Only    few   works    consider    the   problem    of   symmetrical    objects:
Sundermeyer~\textit{et al.}~\cite{Sundermeyer18} solves this problem by learning
a  mapping  to   a  latent  representation  of   the  pose;  Bregier~\textit{et
  al.}~\cite{Bregier:hal-01415027} introduced a representation  of the pose that
differs from rigid motions and suitable  for their similarity metric between two
poses; \cite{Manhardt18}  learns to predict several  poses so that at  least one
pose corresponds to the ground truth; Rad and Lepetit~\cite{Rad17} rely on image
mirroring to deal with some  symmetries.  While these papers propose interesting
solutions, here,  we consider a  general analytic  approach to the  problem.  It
will give insights on the learning-based  methods, and yields a simple method to
solve the ambiguities due to symmetries.

In the remainder of the paper, we  review the state-of-the-art on 3D object pose
estimation  from images,  describe our  method, and  evaluate it  on the  T-Less
dataset, which is made of very challenging objects and sequences.

\section{Related Work}

6-DoF pose estimation made significant progress recently. We discuss
below mostly the most recent ones, and several techniques that have
been proposed to specifically tackle objects with pose ambiguities.

\subsection{6-DoF Object Pose Estimation}
Several  recent works  extend  on  deep architectures  developed  for 2D  object
detection by also predicting the 3D  pose of objects.  \cite{Kehl17} trained the
SSD architecture~\cite{Liu16} to  also predict the 3D rotations  of the objects,
and   the   depths   of   the  objects.    Deep-6DPose~\cite{Do18}   relies   on
Mask-RCNN~\cite{He17}  instead  of  SSD.    To  improve  robustness  to  partial
occlusions, PoseCNN~\cite{Xiang18} segments the  objects' masks and predicts the
objects'  poses   in  the  form  of   a  3D  translation  and   a  3D  rotation.
Yolo6D~\cite{Tekin2018} relies  on Yolo~\cite{Redmon16} and predicts  the object
poses in the form of the 2D projections of the corners of the 3D bounding boxes,
a 3D pose representation introduced in \cite{Crivellaro18} and \cite{Rad17}.

Several   works   also   attempted   to    be   more   robust   to   occlusions.
\cite{Jafari2018,Brachmann16,zakharov2019dpod} first predict  the 3D coordinates
of the  image locations lying on  the objects, in the  object coordinate system,
and  predict the  3D object  pose  through hypotheses  sampling with  preemptive
RANSAC.   \cite{Oberweger18,Peng19}  predict  the 2D  projections  of  3D
points from image  patches or local features, to avoid  the effects of occluders
when performing the prediction.

These works  have been  very successful  at predicting the  3D pose  of objects,
however they  mostly do not  consider objects with  symmetry.  Our goal  in this
paper is  not to  propose another  architecture for 3D  pose prediction,  but to
study the effects of symmetries on the prediction process, and propose a general
solution, which can be integrated in these previous works.

\subsection{Ambiguity Aware Pose Estimation}

\cite{Rad17}  is  probably the  first  work  that  mentioned the  difficulty  of
predicting  the 3D  pose of  objects with  symmetries using  Deep Networks,  and
presents some results on the T-Less dataset. However, the paper does not provide
many details about the method and the  solution is not general.  Our approach is
related to the direction they point at,  but we provide a general solution, with
much more justifications.

\cite{Sundermeyer18} learns a  latent representation of the object  pose using an
auto-encoder.  They show that their  learned embedding is ambiguity agnostic, in
the sense that visually ambiguous poses will  map to the same code in the latent
space. They perform pose estimation by  matching the code obtained from an image
of the object  with a precomputed code  table covering the 6D  pose space. While
this approach is very interesting, we consider here an orthogonal approach based
on an analytical study of the  ambiguities.  Moreover, the code table introduces
some discretization,  while we predict a  3D pose that varies  continuously with
the input image.

\cite{Corona18}  learns  to  compare  an  input image  with  a  set  of
renderings of  the object under many  views, to predict the  most similar view
and to predict the rotational symmetries of the object.  This also requires to
discretize the possible rotations, while we predict a continuous 3D pose.

\cite{Manhardt19} also considers a  learning-based approach, tackles ambiguities
raised by partial  occlusions in addition to  rotational symmetries, \emph{i.e.}
when an  occluder hides  a part  of an  object, so  that it  is not  possible to
estimate the pose exactly anymore. This is done by training a network to predict
multiple poses, so that only one has  to correspond to the actual pose.  At test
time, the network  predicts multiple poses, which are expected  to represent the
distribution  over the  possible poses.   By contrast  with this  learning-based
approach,  we  explicitly   consider  the  ambiguities  that   can  raise  under
symmetries.

\cite{Bregier:hal-01415027} introduced the concept of proper symmetries group in
a survey that aims to cover ambiguities  and a pose representation specific to a
metric on  3D poses.  We use this  concept to solve  the ambiguities  created by
symmetrical objects.  The paper however  does not consider pose prediction using
regression or machine learning.

\cite{Henderson18bmvc} notices  that symmetries produce  multiple modes
in the distribution  $Q(\theta | \bI)$ over 3D poses  $\theta$. They therefore
enforce a  uniform prior  $P(\theta)$ over  symmetrical poses  to successfully
approximate  $Q$.   However,   they  do  not  explicitly   report  results  on
(quasi)-symmetrical objects such as those of T-Less.

\section{Method}
\label{sec:Method}

We  study below  the effect  of  symmetries on  algorithms aiming  to learn  the
mapping between an  image of an object and  its 6D pose, and we show  how we can
derive a simple  method for handling these symmetries.  In  the next section, we
describe how this method can be integrated within a Faster-RCNN framework.

\subsection{Mapping Ambiguous Rotations}
\label{ssec:normalization_one}

Let's    consider     the    set    $\calM(O)$    already     introduced    in
Eq.~\eqref{eq:calM}. In  practice, the motions  in $\calM$ are usually  in the
form $\bm = [R,\boldsymbol{0}]$ with  $R\in \SO(3)$, i.e.  objects have mostly
rotational    symmetries.    A    translation    component   different    from
$\boldsymbol{0}$ would  correspond to  an object with  translation symmetries,
for example a long building with windows of similar appearances.

We thus first  define the notion  of ambiguous rotations:  We say that  two rotations
$R_1$ and $R_2$ are ambiguous if they result in the same object appearance, i.e.
if $\Render(O, [R_1,T_1]) = \Render(O, [R_2,T_2])$.  This defines an equivalence
relationship $R_1 \sim R_2$.  If $R_1 \sim R_2$, then it is not possible from an
image to distinguish between rotation $R_1$  and $R_2$ when predicting the pose.
Predicting $R_1$, or $R_2$, or any rotation  $R \sim R_1$ is equally good.  This
is in fact the idea behind the ADI metric~\cite{Hinterstoisser12}.

\begin{figure}
  \begin{center}
    \includegraphics[width=0.8\linewidth, height=7em]{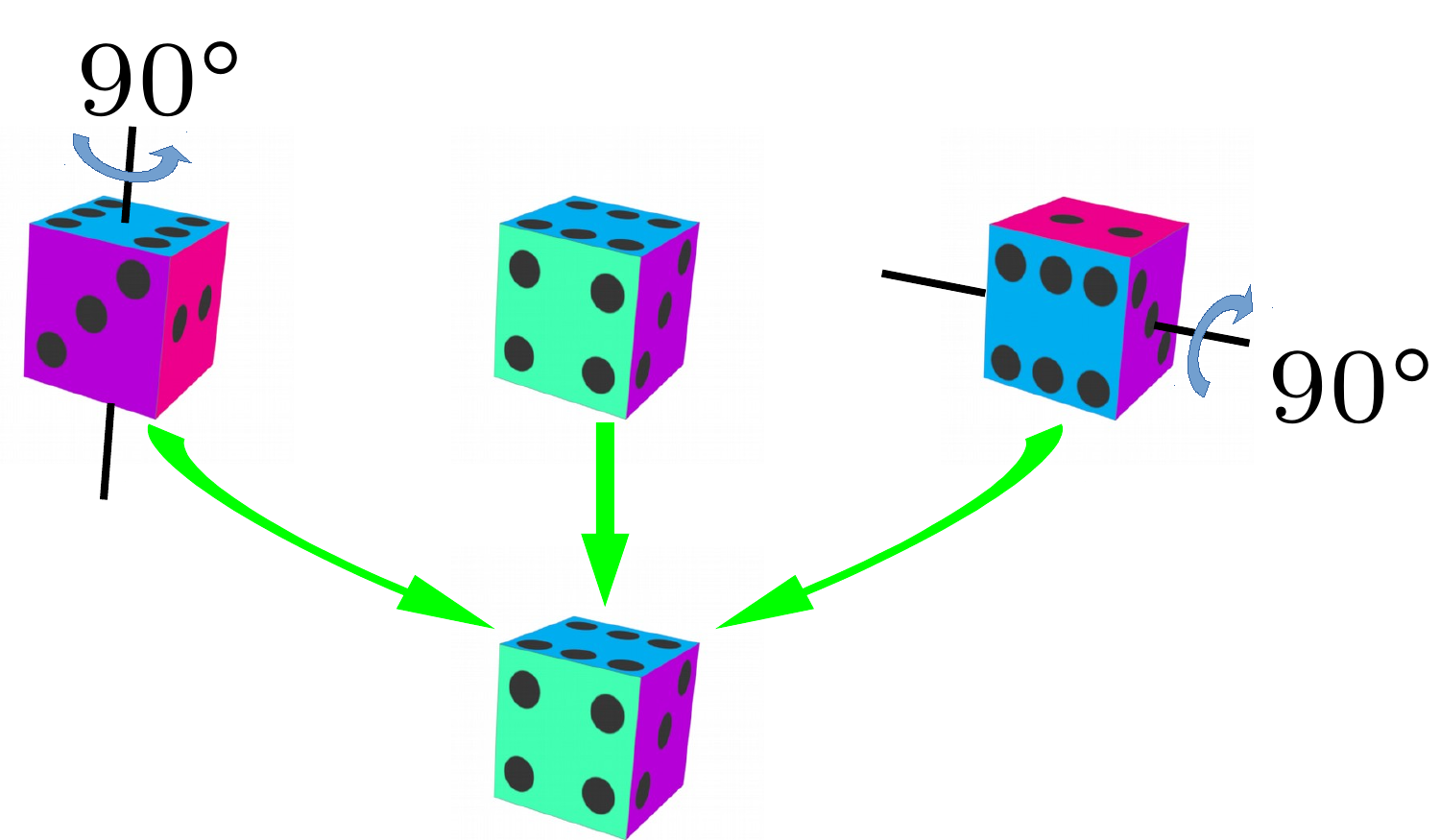}
    \caption{Mapping of 3 ambiguous poses to  the same pose.  We consider here a
      uniform  object  and  the  colors  and  dots on  the  faces  are  only  to
      visualize the  different poses.  The left  and right  poses are
      remapped to the reference pose in the middle.}
    \label{fig:pose_remapping}
  \end{center}
\end{figure}

As  illustrated  in Fig.~\ref{fig:pose_remapping},  a  natural  idea to  aim  at
preventing  trouble  during  learning  is  therefore  to  first  map  equivalent
rotations to a  unique rotation, which we call a  canonical rotation. This means
that during  training, training images with  the same object appearance  will be
assigned  the  same  rotation  after   mapping.   The  transformation  $\calF  :
\bI  \mapsto  \bp$  of  Eq.~\ref{eq:image_to_pose} will  thus  become  a
function and we will  be able to learn it with a Deep  Network for example. This
implies that at inference, the network will predict the canonical rotation for a
given input image, which is the best that can be done in presence of symmetries.

Given set $\calM(O)$ of the object's proper symmetries, we are therefore looking
for an operator $\map(\cdot)$ on $\SO(3)$ that can map ambiguous 3D rotations to
a single rotation such that $\map(R_1) =  \map(R_2) \iff 
R_1 \sim R_2$ ($\star$) holds.

\paragraph{Proposition 1. } \textit{Given a proper symmetry group $\calM(O)$,  
let us define} $\map$ \textit{operator as:}

\begin{equation}
  \map(R) = \hat{S}^{-1}R \> ,\quad \forall R \in \SO(3), 
  \label{eq:map}
\end{equation}
\textit{with} 

\begin{equation}
\hat{S} = \argmin_{S \in \calM(O)} 
\|S^{-1}R - \Id\|_F \> , 
  \label{eq:S_hat}
\end{equation}

\textit{where  $\| .   \|_F$ is  the  Froebenius norm. Then} $\map$ 
\textit{verifies the mapping property ($\star$).}

\paragraph{Proof. }
To simplify  the notations, let us consider that $\calM(O)$ is made 
only of the rotation components.
By definition of $R_1 \sim R_2$ and $\calM(O)$:
\begin{equation}
  R_1 \sim R_2 \Leftrightarrow \exists! \> S_{12} \in \calM(O) \>\>\> 
  \textrm{s.t.} 
  \>\> R_1 = S_{12}R_2 \> . 
\end{equation}
Let us consider the solution of the optimization problem in Eq.~\eqref{eq:S_hat}
for $R_1$:
\begin{equation}
 \hat{S}_1 = \argmin_{S \in \calM(O)} \|S^{-1}R_1 - \Id \|_F \> . 
\end{equation}

then

\begin{equation}
 \hat{S}_1 = \argmin_{S \in \calM(O)} \|S^{-1}S_{12}R_2 - \Id \|_F \> . 
\end{equation}
We introduce variable $T$ such that $S=S_{12}T$. Since $S$ and $S_{12}$ belong to
$\calM(O)$ and $\calM(O)$ is a group, $T$ also belongs to $\calM(O)$. We can therefore
perform the following change of variable:
\begin{equation}
 \hat{S}_1 = S_{12} \argmin_{T \in \calM(O)} \|T^{-1}R_2 - \Id \|_F \> ,
\end{equation}
 which is equal to:
\begin{equation}
 \hat{S}_1 = S_{12} \hat{S}_2 \> , 
\end{equation}
with
\begin{equation}
 \hat{S}_2 = \argmin_{S \in \calM(O)} \|S^{-1}R_2 - \Id \|_F \> . 
\end{equation}
Therefore
\begin{equation}
  \begin{array}{lcl}
  R_1 \sim R_2 &\Leftrightarrow & \map(R_1) = \hat{S}_1^{-1}R_1 = \hat{S}_2^{-1}
  S_{12}^{-1} S_{12} R_2 \\ &&=\hat{S}_2^{-1} R_2 = \map(R_2) \> . 
\end{array} 
\end{equation} \qed

\subsection{Implementing $\map$}

If $\calM$ is discrete, implementing operator $\map$ is trivial, as it is only a
matter of iterating  over the elements of $\calM$ to  find the minimum. However,
$\calM$ can  be continuous for some  objects.  This is the  case for generalized
cylinders and spheres~\cite{Bregier:hal-01415027}.  For  spheres, $\map$ is also
trivial as it can always return the identity transformation, for example.

For generalized  cylinders, implementing  operator $\map$  is more  complex.  In
this case, $\calM$ can be written as:
\begin{equation}
  \calM(O) = \{R^\bu_\alpha : \alpha \in [0, 2\pi)\}, 
\end{equation}
where $R^\bu_\alpha$ is the rotation around axis $\bu$ of amount $\alpha$. 

The Froebenius norm in Eq.~\eqref{eq:S_hat} can be rewritten as
\begin{equation}
  \|S^{-1} R - \Id \|_F = \|D\|_F = \tr{D^T D} \> , 
\end{equation}
with $D = S^{-1} R - \Id$. After some derivations:
\begin{equation}
  \|S^{-1} R - \Id \|_F = 6 - \tr{S^T R} \> .
\end{equation}
The complete derivations can be found in the supplementary material.

Without loss of  generality, we can assume  that $\bu$ is the  $\bz$-axis of the
object's coordinate  system: If it  is not the  case, the following  still holds
after applying a change of basis. Then, $S$ has the following form:
\begin{equation}
  S = \begin{bmatrix}
    \cos(\alpha) & -\sin(\alpha) & 0 \\
    \sin(\alpha) &  \cos(\alpha) & 0 \\
    0 & 0 & 1
  \end{bmatrix} \> , 
\end{equation}
and, after some basic manipulation:
\begin{equation}
  \begin{aligned}
    \tr{D^T D} = 6 &- (R_{11} + R_{22})\Ca \\
    &+ (R_{12} - R_{21})\Sa \> . 
  \end{aligned}
\end{equation}
To  implement $\map$,  we  need to  solve the  optimization  problem $\hat{S}  =
\argmin\limits_{S\in\calM(O)}  \tr{D^T D}$,  which  can now  be  rewritten as  a
minimization over $\alpha$:
\begin{equation}
    \begin{aligned}
      \hat{\alpha} &= \argmin\limits_{\alpha \in [0, 2\pi)} \tr{D^T D} \\
	&= \argmax\limits_{\alpha \in [0, 2\pi)} \>(R_{11} + R_{22})\Ca - 
	  (R_{12} - R_{21})\Sa \> . 
    \end{aligned}
\end{equation}
This is solved analytically by solving $\frac{\partial\tr{D^T D}}{\partial 
  \alpha}=0$ for $\alpha$. The solution of Eq.~\eqref{eq:S_hat} is then:
\begin{equation}
  \hat{S} = R^\bz_{\hat{\alpha}} \quad \textrm{with } \>\> \hat{\alpha} = 
  \text{arctan2}(R_{21} - R_{12}, R_{11} + R_{22}) \> . 
  \label{eq:atan2}
\end{equation}

\subsection{Discontinuities of $\calF$ After Mapping}
\label{ssec:normalization_two}

After  applying the  $\map$ operator,  there  are no  pose ambiguities  anymore,
\emph{i.e.}  two similar images are assigned  the same rotation.  However, a new
difficulty  arises:   The  transformation   $\calF(\bI)\rightarrow  \bp$   is  
now discontinuous  around  some rotations.   This  is  problematic when  using  
Deep Networks  to learn  $\calF$, as  Deep Networks  can only  approximate 
continuous functions~\cite{Cybenko89, Hornik91,Hanin17}.

\begin{figure}
  \begin{center}
    \icg{0.3}{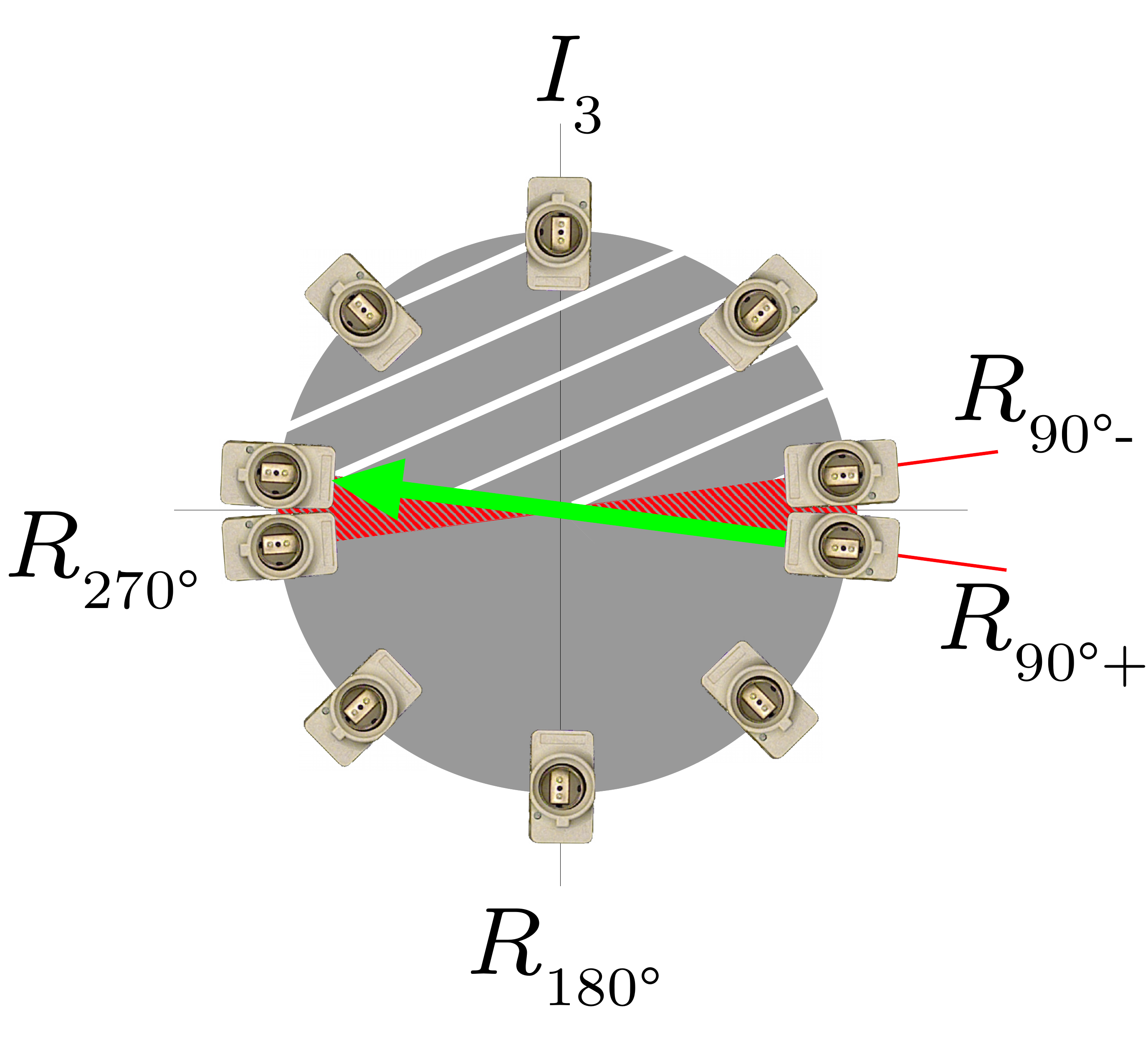}
    \caption{Discontinuities of  $\calF$ after  applying the  $\map$ operator,
      for an object with one axis of symmetry and a $\pi$-symmetry.  All poses
      are mapped to a pose in  the hashed region by operator $\map$ introduced
      in   Section~\ref{ssec:normalization_one}.   Since   $\map(R^z_{\pi/2  +
        \epsilon}) = R^z_{\epsilon  - \pi/2}$ (visualized by  the green arrow)
      and  $\map(R^z_{\pi/2  - \epsilon})  =  R^z_{\pi/2  - \epsilon}$,  there
      exists a hazardous region (in red) where $\calF$ is
      discontinuous.}
    \label{fig:discontinuities_problem}
  \end{center}
\end{figure}

To understand why these discontinuities happen, let us consider an example, more
exactly    the    rectangular    object    seen   from    the    top    as    in
Fig.~\ref{fig:discontinuities_problem}.   $\calM(O)$ is  made  of two  rotations
around the $Z$ axis: The identity matrix,  and the rotation of angle $\pi$, and
 $\calM(O)=\{\Id,  R^\bu_\pi\}$.   If  a training  image  is  annotated  with
rotation  $R_{\pi/2+\epsilon}^\bz$, this  rotation  will be  mapped by  operator
$\map$ to rotation $R_{\epsilon - \pi/2}^\bz$;  If a training image is annotated
with  pose $R_{\pi/2-\epsilon}^\bz$,  this  rotation will  be  mapped to  itself
\emph{i.e.}  $R_{\pi/2-\epsilon}^\bz$.   By making $\epsilon$ converge  to 0, it
can be  seen that there  is a discontinuity  of $\calF$ around  images annotated
with rotations $\pi$ before mapping.

Another way  of looking at the  problem is to  notice that images of  the object
annotated with rotations $R_{\epsilon - \pi/2}^\bz$ and $R_{\pi/2-\epsilon}^\bz$
look very similar, but with very different rotations.  A Deep Network would have
to learn to predict very different poses for very similar images.

\subsection{Solving the Discontinuities}
\label{sec:solving_disc}

The discontinuities  only occur when  $\calM$ is discrete:  It can be  seen from
Eq.~\eqref{eq:atan2}  that in  the case  of a  generalized cylinder,  the $\map$
operator  is   continuous.   Otherwise,   we  avoid  these   discontinuities  by
introducing a  partition of $\SO(3)$ made  of two subsets $\W_1$  and $\W_2$.  For
each subset, we train a different regressor to predict the pose. We will therefore
have two regressors  $\calF_1$ and $\calF_2$ instead of only  one.  In this way,
both $\calF_1$ and $\calF_2$ will be continuous over their respective domains.

We describe below our method on an example, and then extend it to the general case.

\subsubsection{One Symmetry Axis, $M = 2$}\label{sssec:ru_2}

Let    us    consider    again    the    rectangular    object    pictured    in
Fig.~\ref{fig:discontinuities_problem},     and     already     discussed     in
Section~\ref{ssec:normalization_two}. For this object, we have $\calM(O) =
\{\Id, R^\bu_\pi\}$. We can notice that $\calM$ and $\map$ generate a partition
of $\SO(3)$ made of two subsets:
\begin{equation}
  \W_1 = \{ R : \hat{S}(R) = \Id\} \text{ and } \W_2 = \{ R : \hat{S}(R) = 
  R^\bu_\pi\} \> , 
  \label{eq:first_W1_W2}
\end{equation}
where $\hat{S}(R)$ is the rotation  of Eq.~\eqref{eq:S_hat} when applying $\map$
to $R$.

\begin{figure}
  \begin{center}
    \begin{tabular}{cc}
      \begin{subfigure}[t]{0.45\linewidth}
        \includegraphics[width=\linewidth]{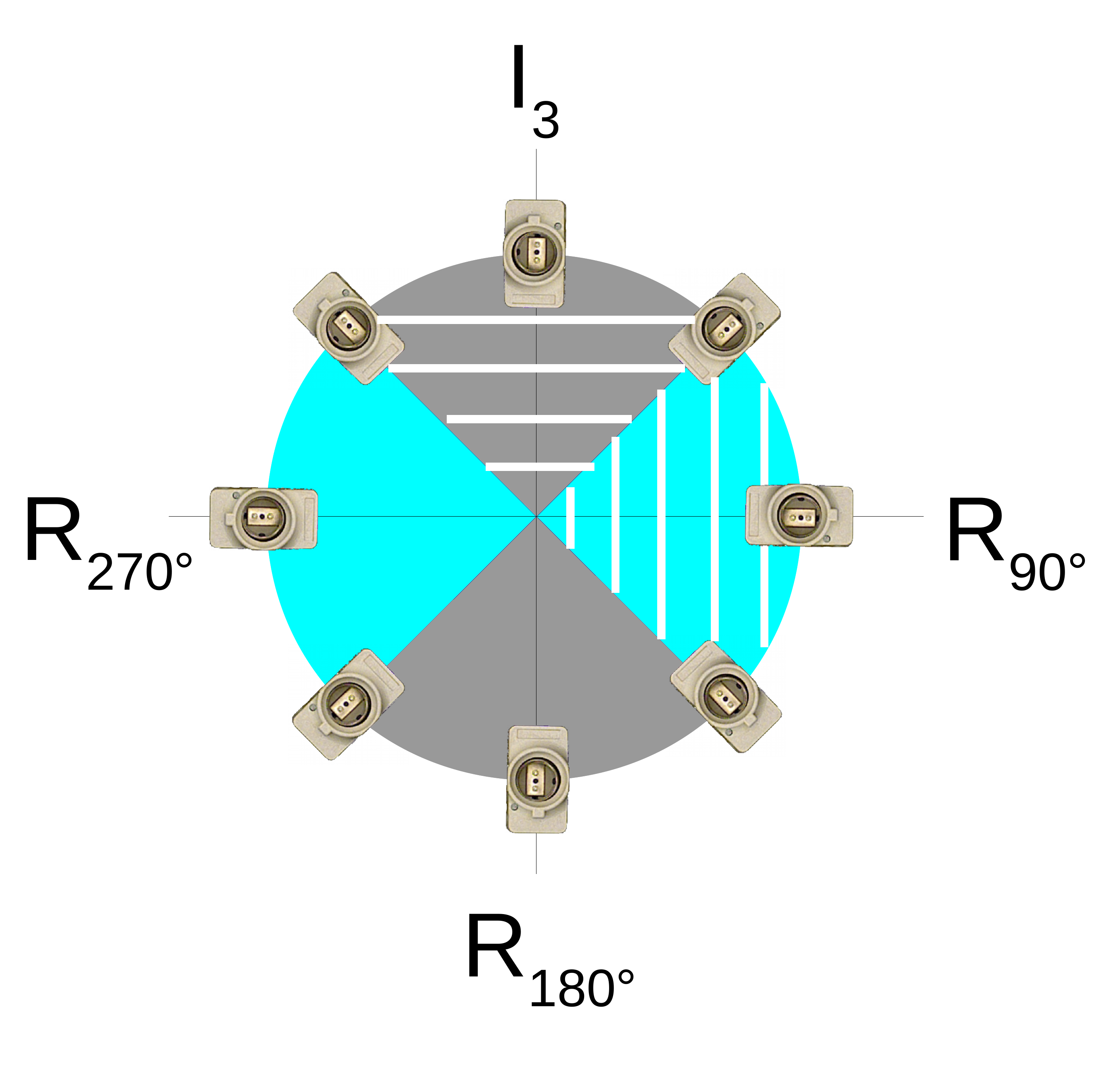}
        \caption{}
        \label{fig:normalization_180}
      \end{subfigure} &
      \begin{subfigure}[t]{0.45\linewidth}
        \includegraphics[width=\linewidth]{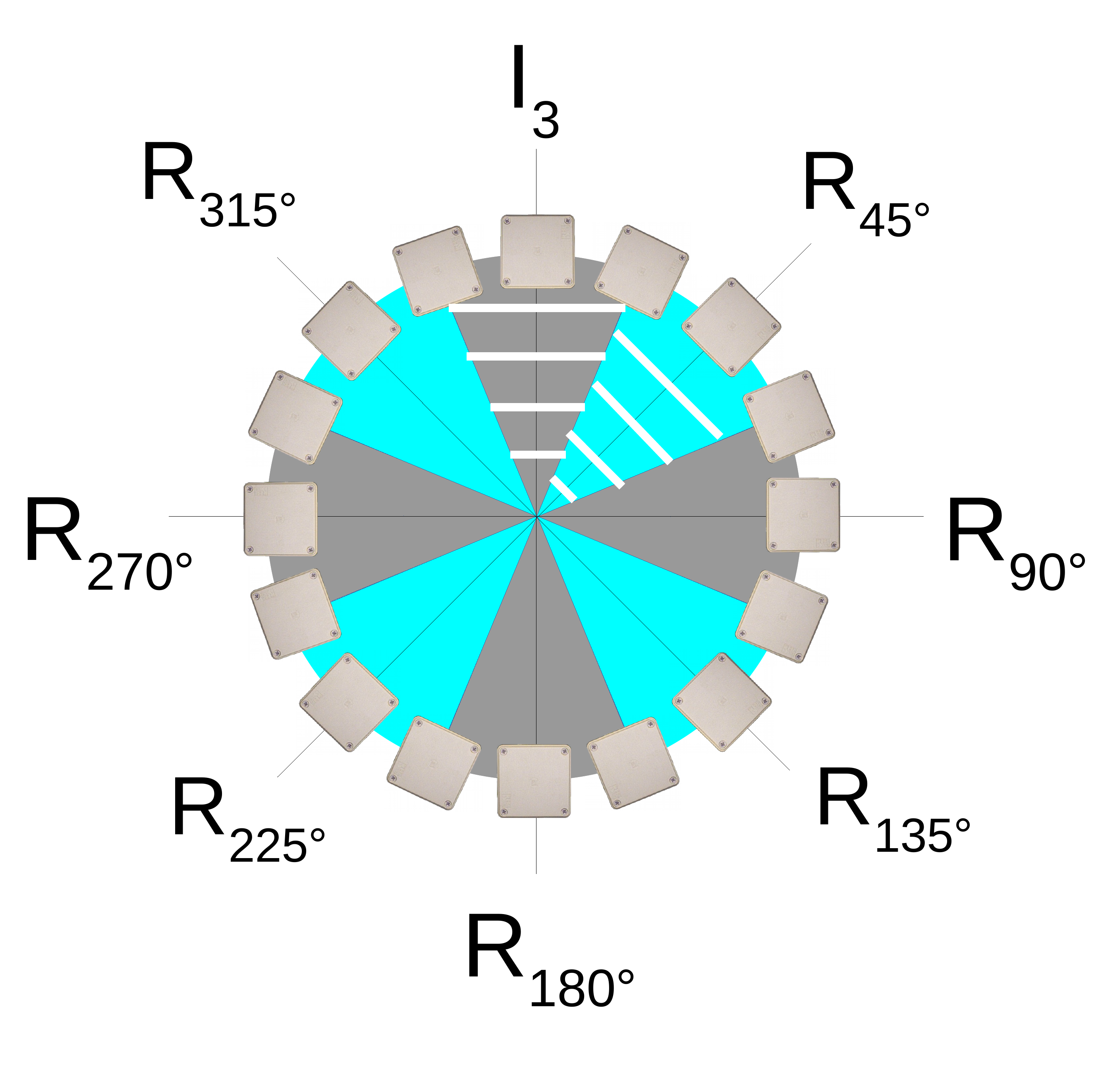}
        \caption{}
        \label{fig:normalization_90}
      \end{subfigure}
    \end{tabular}
  \end{center}
  \caption{\label{fig:discontinuities_solving} Partitions  for an  object with
    one  axis of  symmetry with  $M=2$  (left) and  $M=4$ (right)  as defined  in
    Section~\ref{sec:solving_disc}. Rotations in areas  filled with one color  should be
    mapped  to a rotation in the  hashed  region of same color  to  avoid  
    discontinuities.  Two different regressors $\calF_1$ and $\calF_2$, one for
    each color, are used to predict poses for each hashed region.}
\end{figure}

However, this partition will not solve our problem: We already know that $\calF$
is not  continuous on $\W_1$.   We must therefore  introduce a new  partition of
$\SO(3)$. For this partition, we consider the new set:
\begin{equation}
  \begin{aligned}
    \sqrt{\calM}(O) &= \{(R^\bu_{k\pi/2}) : k \in \mathbb{Z}\} \\
    &= \{\Id, R^\bu_{\pi/2}, R^\bu_{\pi},R^\bu_{3\pi/2}\} \> ,
  \end{aligned}
  \label{eq:calMprime_2}
\end{equation}
and the partition it generates with $\map$:
\begin{equation}
  \W^{(k)} = \{ R : \hat{S}(R) = R^\bu_{k\pi/2}\} \> . 
\end{equation}
As shown  in Fig.~\ref{fig:normalization_90}, no part $\W^{(k)}$ include
any discontinuity. Moreover, for a rotation in $\W^{(2)}$, there is another
rotation in $\W^{(0)}$ that generates the same object appearance. The same
yields for $\W^{(3)}$ and $\W^{(1)}$.

We therefore take  $\W_1 = \W^{(0)}$ for the domain  of regressor $\calF_1$, and
$\W_2  =  \W^{(1)}$  for  the  domain of  regressor  $\calF_2$.   $\calF_1$  and
$\calF_2$  thus do  not suffer  from  discontinuities nor  ambiguity.  They  are
sufficient to estimate the object pose under any rotation, since we can map this
rotation to a rotation  either in $\W_1$ or $\W_2$ corresponding to the same
appearance.  To  do so, we introduce  a new mapping $\map'$  derived from $\map$
such that:

\begin{equation}
	\begin{aligned}
	&\forall R\in SO(3), \quad \map'(R) = (\hat{S}^{-1} R, \delta) \>\> 
	\textrm{such that } (\hat{S}, \delta) = \\
	&\textrm{\begin{tabular}{p{0.95\linewidth}}  
		$\left\{
		\textrm{\begin{tabular}{p{1\linewidth}}
			$(\argmin\limits_{S\in\calM(O)}  \|S^{-1}R - \Id \|_F , 1) 
			\quad$ if 
			$\map(R)\in \W_1$  ,\\
			$(\argmin\limits_{S\in\calM(O)}  \|S^{-1}R - R^\bu_{\pi/2} \|_F,
			2) \quad$ otherwise , 
			\end{tabular}}
		\right.$
		\end{tabular}}
	\end{aligned}
\label{eq:rotation_remapping_two_2}
\end{equation} 

During training,  given a  training image  $\bI$ annotated  with rotation  $R$, we
compute  $(\hat{S}^{-1}  R, \delta)  \leftarrow  \map'(R)$  and train  regressor
$\calF_\delta$ to predict rotation $\hat{S}^{-1} R$ from $\bI$.

During inference,  given a test image  $\bI$ of an  object, we need to  know which
regressor we should  invoke to predict the  pose. To do so,  during training, we
train  a classifier  $\calC$  to predict  which regressor  we  should invoke  to
compute the  pose, that is  we train $\calC$ to  predict $\delta$ from  $\bI$. For
rotations close  to the boundary between  $\W_1$ and $\W_2$, the  prediction for
$\calC$ can  become ambiguous.  However,  in this case,  the ambiguity is  not a
problem in practice: Even if the  classifier predicts the wrong regressor to use
close to the  boundary between $\W_1$ and $\W_2$, both  regressors can correctly
predict poses close to this boundary.

\subsubsection{One Symmetry Axis, Arbitrary $M$}

Let us now generalize to an object $O$ with an arbitrary amount of symmetries around a single axis $u$.  These symmetries are necessarily periodic around $\bu$ with angular period $f_\alpha = 2\pi / M$: Rotating $O$ around $u$ by any angle multiple of $f_\alpha$ does not change its appearance.  The proper symmetry group $\calM(O)$ for such an object is:
\begin{equation}
  \calM(O) = \left\{\left(R^\bu_{2\pi/M}\right)^m\right\}_{m\in \IN} =
  \{R^\bu_{2m\pi/M}\}_{m\in \IN} \> . 
  \label{eq:calM_M}
\end{equation}	
$\sqrt{\calM}(O)$ of Eq.~\eqref{eq:calMprime_2}) becomes:
\begin{equation}
  \sqrt{\calM}(O) = \left\{\left(R^\bu_{\pi/M}\right)^m\right\}_{k\in \IN} = \{R^\bu_{m\pi/M}\}_{m\in \IN} \> , 
\end{equation}	
and mapping $\map'$ of Eq.~\eqref{eq:rotation_remapping_two_2} becomes:
\begin{equation}
    \begin{aligned}
      &\forall R\in SO(3), \quad \map'(R) = (\hat{S}^{-1} R, \delta) \>\> 
      \textrm{such that } (\hat{S}, \delta) = \\
      &\textrm{\begin{tabular}{p{0.95\linewidth}}  
          $\left\{
          \textrm{\begin{tabular}{p{1\linewidth}}
              $(\argmin\limits_{S\in\calM(O)}  \|S^{-1}R - \Id \|_F , 1) 
              \quad$ if 
              $\map(R)\in \W_1$  ,\\
              $(\argmin\limits_{S\in\calM(O)}  \|S^{-1}R - R^\bu_{\pi/M} \|_F,
              2) \quad$ otherwise , 
          \end{tabular}}
          \right.$
      \end{tabular}}
    \end{aligned}
  \label{eq:rotation_remapping_one_axis}
\end{equation} 

where $\W_1 = \{ R : \hat{S}(R) = \Id\}$.

We can use $\map'$  the same way as in the previous subsection  to train and use to
regressors $\calF_1$ and $\calF_2$.

\subsubsection{General Case}

In the general case, each rotation $R$ in $\calM$ can be written in the form:
\begin{equation}
  R = R_{2\pi/M}^\bu . R_{2\pi/N}^\bv .  ...  \quad \textrm{with } M, N, .. \in
  \IN \> , 
\end{equation}
where $\bu$, $\bv$, etc.  are rotation axes. Most common objects  have at most 2
axes of symmetries, but it is possible to imagine objects with more, for example
a golf ball. To keep the notations as  simple as possible, we will stick to only
two axes, as it is easy to extend to more axes from there.

$\sqrt{\calM}(O)$ becomes:
\begin{equation}
  \sqrt{\calM}(O) = \{ R^\bu_{m\pi/M} . R^\bu_{n\pi/N} \}_{(m, n)\in \IN^2} \> , 
  \label{eq:calMprime_M}
\end{equation}	
and mapping $\map'$ becomes:
\begin{equation}
    \begin{aligned}
      &\forall R\in SO(3), \map'(R) = (\hat{S}^{-1} R, \delta_1,\delta_2) 
      \>\> \textrm{s.t. }  (\hat{S}, \delta_1,\delta_2) = \\
      &\textrm{\begin{tabular}{p{0.95\linewidth}}  
          $ \left\{
          \textrm{\begin{tabular}{p{1\linewidth}}
              $(\argmin\limits_{S\in\calM(O)}  \|S^{-1}R - \Id \|_F , 1, 1)$ if $\map(R)\in \W_{1,1}$  ,\\
              $(\argmin\limits_{S\in\calM(O)}  \|S^{-1}R - R^\bu_{\pi/M} \|_F, 2,1)$ if $\map(R)\in \W_{2,1}$  ,\\
              $(\argmin\limits_{S\in\calM(O)}  \|S^{-1}R - R^\bv_{\pi/N} \|_F, 1,2)$ if $\map(R)\in \W_{1,2}$  ,\\
              $(\argmin\limits_{S\in\calM(O)}  \|S^{-1}R - R^\bu_{\pi/M}R^\bv_{\pi/N} \|_F, 2,2)$ otherwise , 
          \end{tabular}}
          \right.$
      \end{tabular}}
    \end{aligned}
  \label{eq:rotation_remapping_two_axes}
\end{equation} 
where $\W_{1,1} = \{ R : \hat{S}(R)  = \Id\}$, $\W_{2,1} = \{ R : \hat{S}(R)
= R^\bu_{\pi/M}\}$,  and $\W_{1,2} = \{  R : \hat{S}(R) =  R^\bv_{\pi/N}\}$.  It
means that in this case, we have  to train 4 different regressors $\calF_{1,1}$,
$\calF_{2,1}$, $\calF_{1,2}$, and $\calF_{2,2}$ according to $\delta_1$ and $\delta_2$, and
the classifier $\calC$ to predict a class index in $[0;3]$.

\subsection{Method Summary}

The method developed above can be summarized as follow. We distinguish between 
generalized cylinders and objects with discrete symmetries.

If the object is a generalized  cylinder, given a training image $\bI$ annotated
with rotation  $R$, we  train a  single regressor  $\calF$ to  predict $\map(R)$
using Eq.~\ref{eq:map} from $\bI$. At inference  time, given a test image $\bI$,
we simply have to invoke $\calF$ to predict the object pose from $\bI$.

If the  object has discrete symmetries,  given a training image  $\bI$ annotated
with     rotation     $R$,     we     apply     $\map'$     to     $R$     using
Eq.~\eqref{eq:rotation_remapping_one_axis}                                    or
Eq.~\eqref{eq:rotation_remapping_two_axes} depending  on the number  of symmetry
axes.  $\map'$ provides  the rotation to be associated with  $\bI$ for training,
as  well as  the index  of the  regressor $\calF_i$  to train.   In addition  to
training the  regressors, we  need to  train classifier  $\calC$ to  predict the
index of  the regressor to use.   At inference time, we  first invoke classifier
$\calC$ to  predict which regressor we  should use from $\bI$,  and then, invoke
this regressor to predict the object pose from $\bI$.

\section{Integration into Faster-RCNN}

We integrated our approach into  Faster-RCNN~\cite{Ren15}.  We keep the original
architecture of~\cite{Ren15}  to obtain  region proposals  and classify  each of
those regions with an object  label: In the T-Less dataset~\cite{Hodan17}, there
exist 30 classes of object. We also keep the original loss terms for this part.

We chose to predict the objects' 6D poses in the form of the 2D reprojections of
the      8      corners     of      the      3D      bounding     boxes,      as
in~\cite{Rad17,Tekin2018,Tremblay18,Peng19}   for   simplicity.    From   these   2D
reprojections,  it   is  possible  to   estimate  a   6D  pose  using   a  P$n$P
algorithm~\cite{Hartley00}.   However, our  approach is  general, and  using any
other  representation  of  the  pose,  with quaternions  for  example,  is  also
possible.

\begin{figure}[t]
\begin{center}
  \icg{0.42}{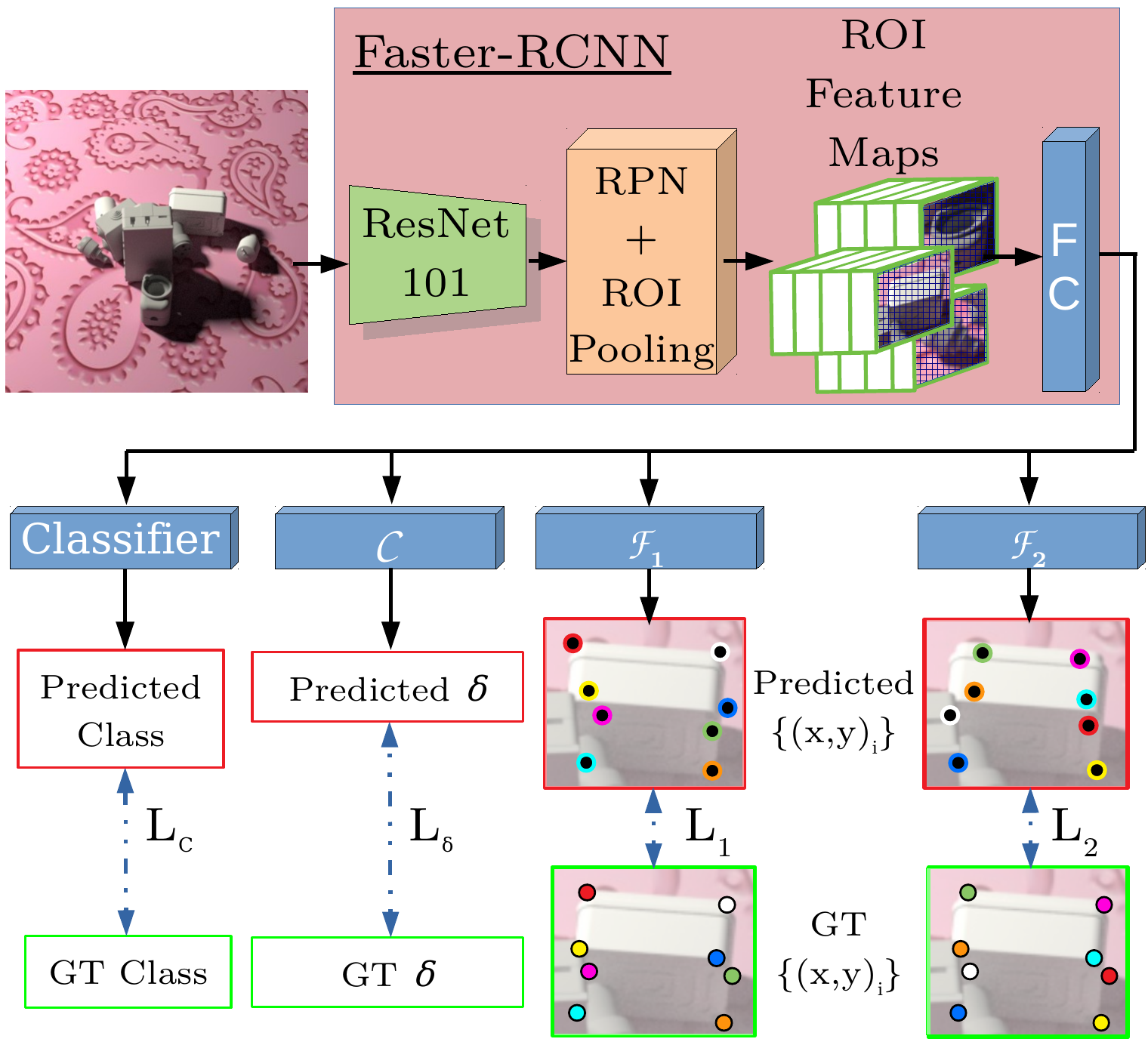}
  \vspace{-0.6cm}
  \caption{Our architecture for implementing our  approach. It is built on top
    of  the Faster-RCNN~\cite{Ren15}  architecture, to  which we  add specific
    branches: One for each regressor $\calF_i$, and one for classifier $\calC$
    to learn to choose between the regressors.}
  \label{fig:Faster-RCNN}
\end{center}

\end{figure}

Fig.~\ref{fig:Faster-RCNN} shows the different branches we added to the original
Faster-RCNN architecture. We describe them below.

\vspace{-0.2cm}
\paragraph{Pose regressor $\calF$ branch.}
We add a specific branch to the Faster-RCNN~\cite{Ren15} architecture to predict
the 2D  coordinates of each  3D corner for each  regressor.  The output  of each
branch has size $16  \times 30$, where $30$ is the number  of object classes and
$16$ accounts for  the 8 2D coordinates to predict.   This branch is implemented
as a fully connected multi-layer perceptron and takes as input the output shared
single  channel  feature-map.  We  then  use  an $L_1$  or  $L_2$  loss on  each
coordinate.  More details can be found in the supplementary materials.

\vspace{-0.2cm}
\paragraph{Classifier $\calC$ branch.}
We  also  added a  specific  multi-layer  perceptron  branch to  Faster-RCNN  to
implement    classifier   $\calC$.     Ground    truth    is   obtained    using
Eq.~\eqref{eq:rotation_remapping_two_2}.

\section{Experiments}

In  this  section,  we detail  how  we  evaluated  our  approach, and  show  its
effectiveness on objects with various types of symmetries.

\subsection{Dataset}

We use  the objects of the  T-Less dataset~\cite{Hodan17} as they  exhibit many
different challenges  due to  symmetries, and are  representative of  objects in
daily and industrial environments.  However, the T-Less dataset does not provide
many  images for  training,  with  a limited  range  of  poses and  illumination
conditions.   We therefore  generated training  and  test images  using the  CAD
models  provided with  T-Less, introducing  partial occlusions  and illumination
variations. This dataset is made of  30K samples, generated using the CAD models
provided in the original T-Less  dataset with Cycles, a photorealistic rendering
engine of the open source software Blender.

\begin{figure}[t]
  \begin{center}
      \begin{tabular}{ccc}
	\icg{0.12}{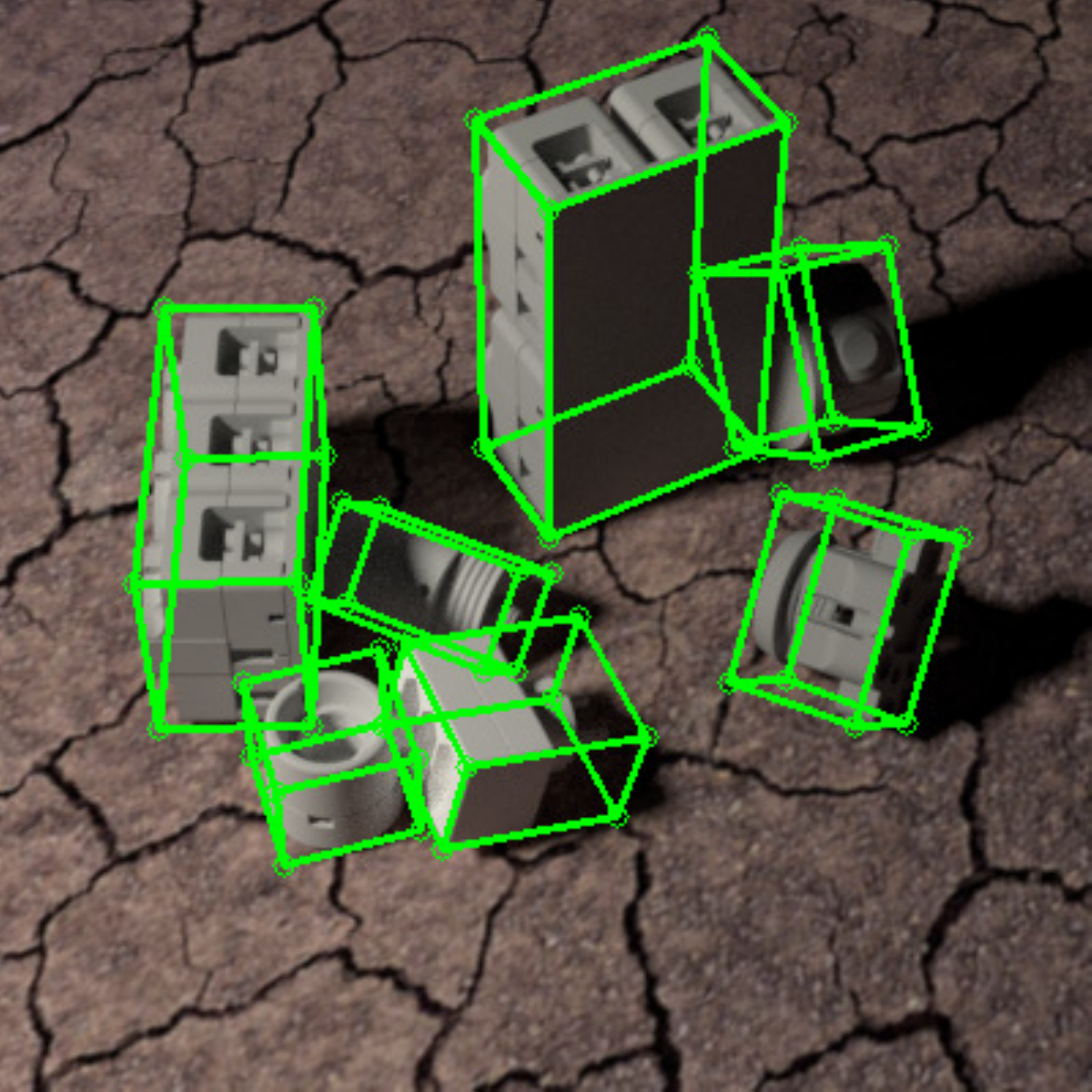} & 
	\icg{0.12}{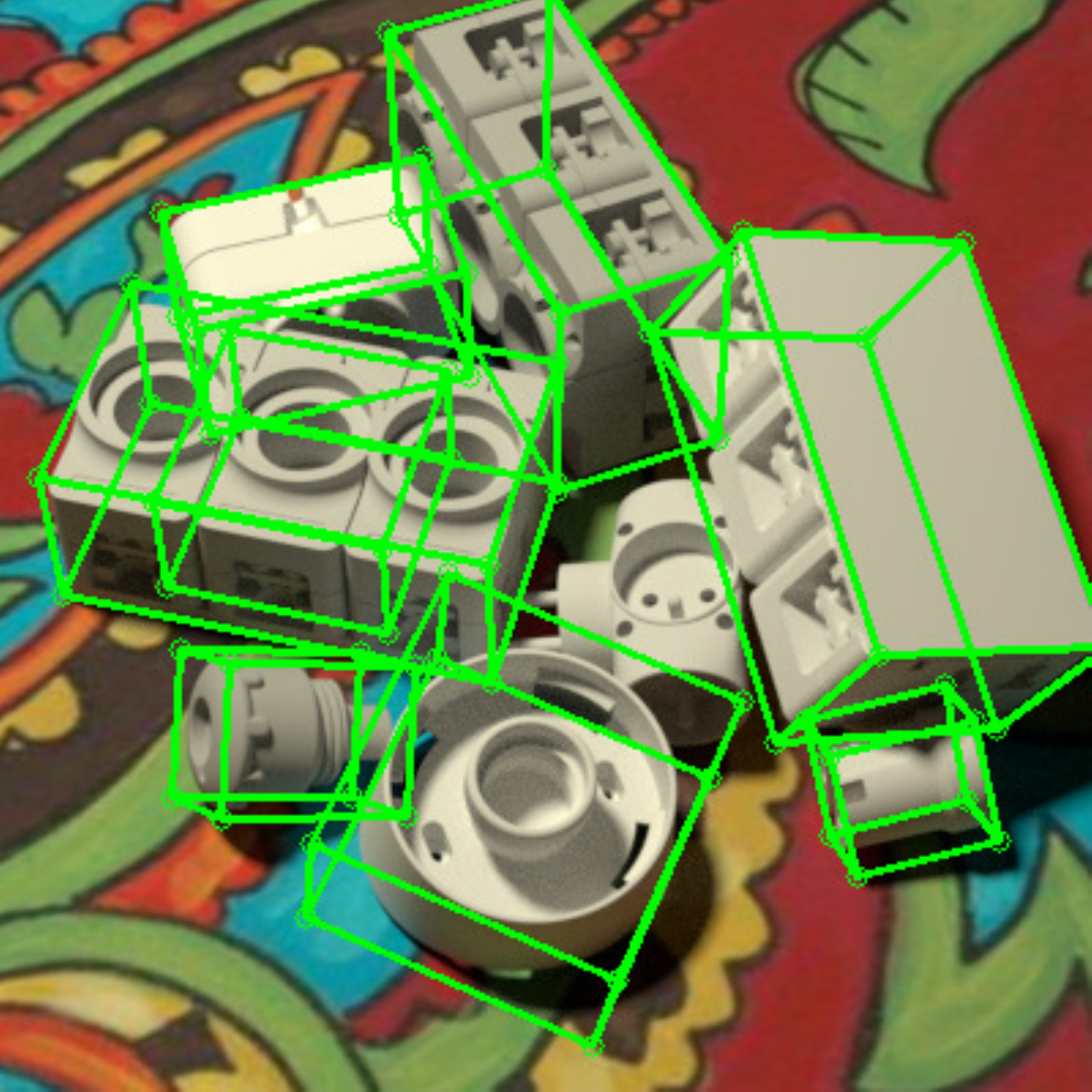} & 
	\icg{0.12}{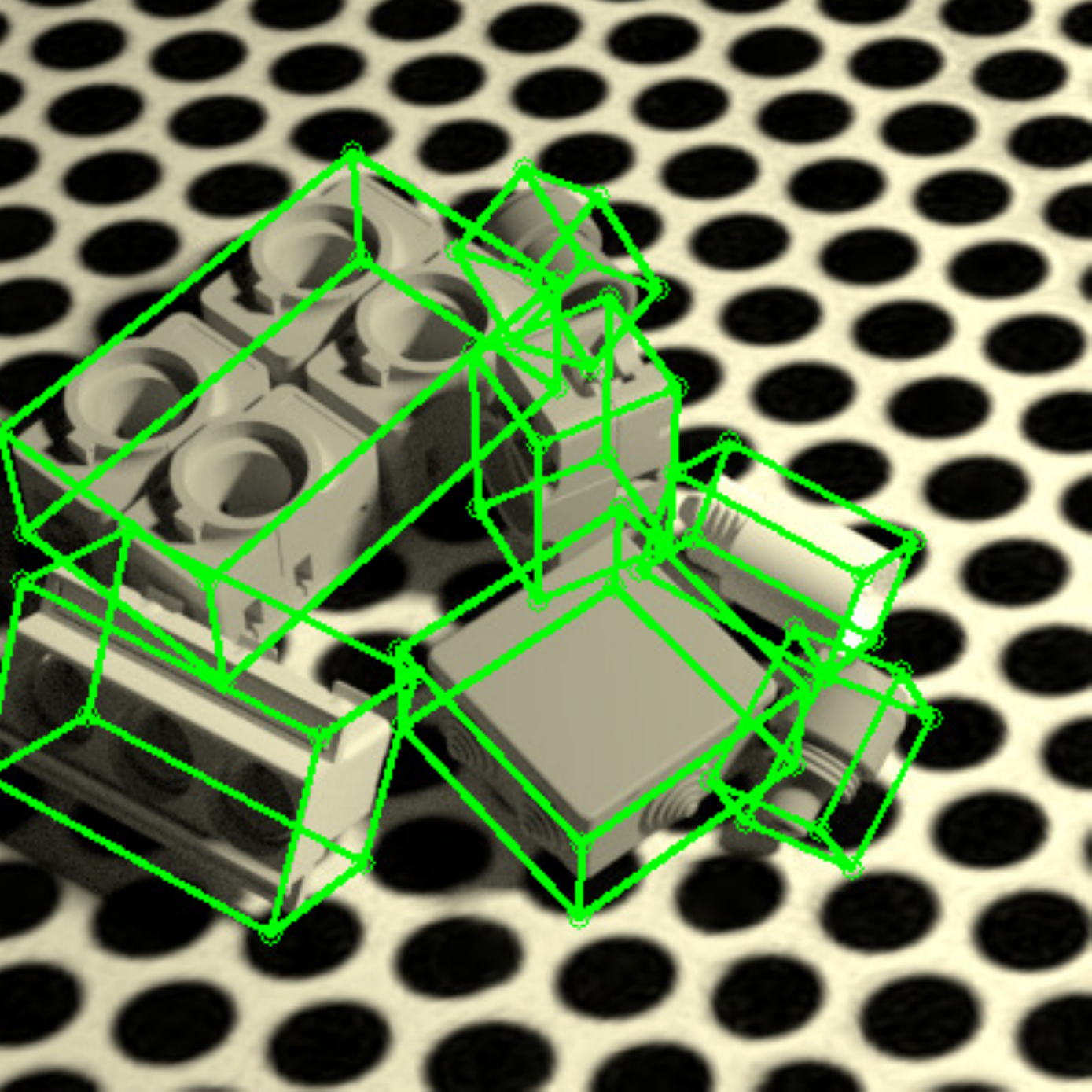} \\
	\icg{0.12}{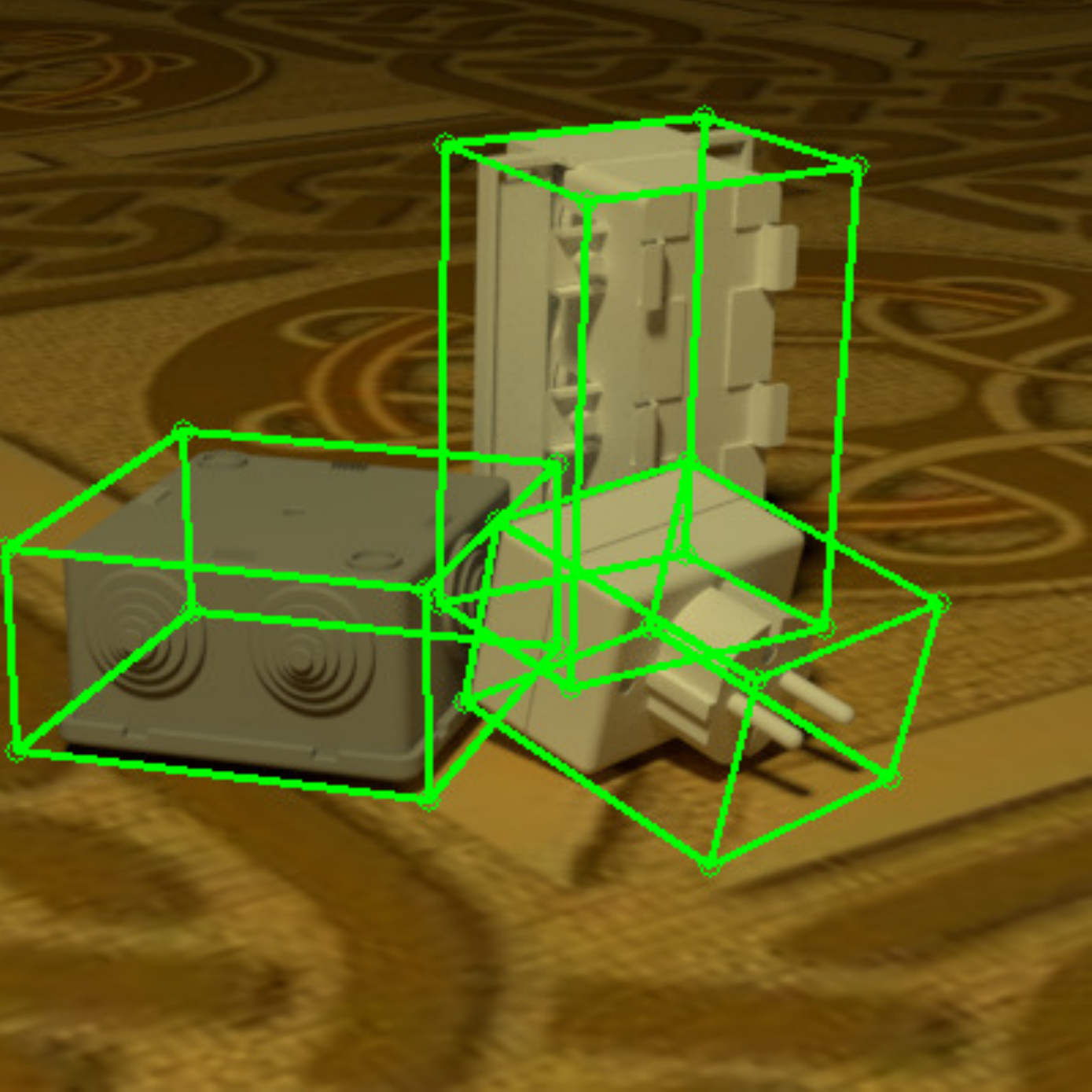} &
	\icg{0.12}{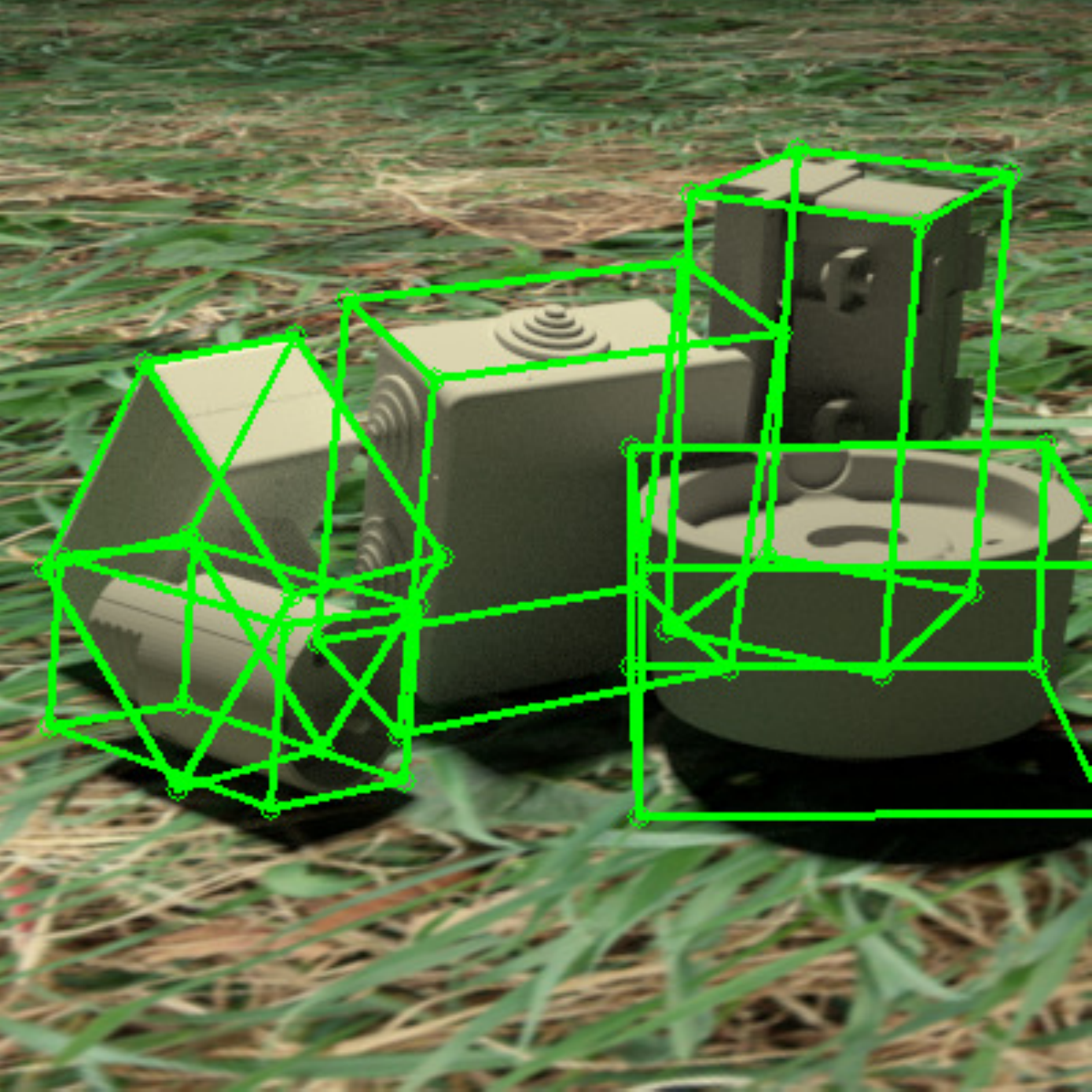} &
	\icg{0.12}{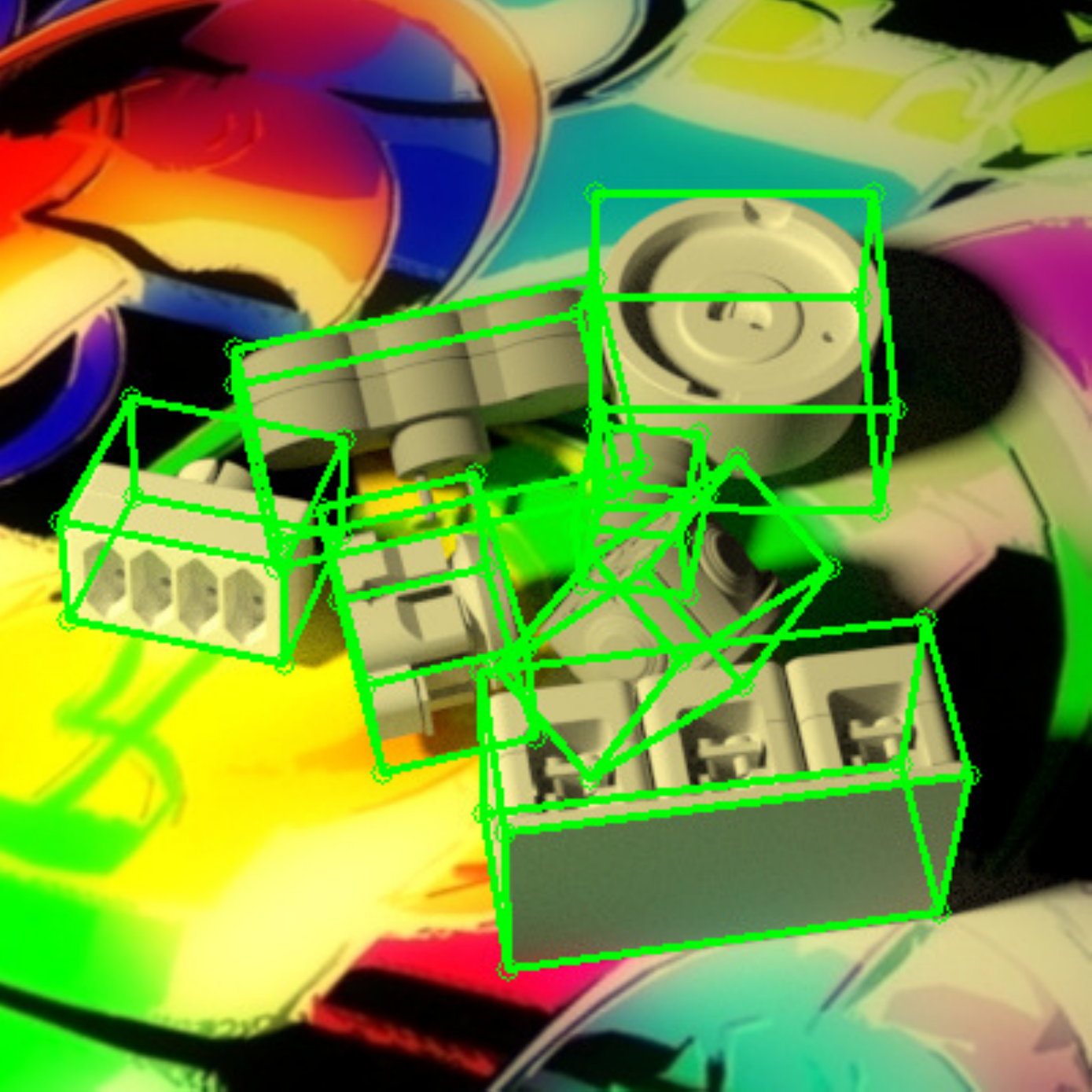}  
    \end{tabular}
    \vspace{-0.6cm}
    \caption{Sample images from our SyntheT-Less dataset. All objects in each image 
      are annotated with their classes and 6D poses.}
    \label{fig:synthetless_sample}
  \end{center}
\end{figure}

Each sample  of our  dataset is  generated using a  random set  $\mathcal{S}$ of
objects  taken from  the T-Less  dataset, using  random gray  scale color  (from
dark-gray to white) for each of  them. Each object of $\mathcal{S}$ is initially
set with a random pose, and we let  the objects fall down on a randomly textured
plane, using Blender's  physics simulator. Because the  objects can collide 
together,   their   final  pose   on  the   table  is   also random. 
Illumination randomization is performed  by varying the level of ambient
light and randomizing  a point light source in terms  of position, strength, and
color.   This  often  results  in  strong  cast  shadows,  as  can  be  seen  in
Fig.~\ref{fig:synthetless_sample}.   A  comparison   with  the  original  T-Less
(primesense) dataset is given in Table~\ref{tab:compare_datasets}.  This dataset
therefore  provides  challenging conditions,  and  allows  us  to focus  on  the
challenges raised by  the symmetries, without having to consider  the domain gap
between synthetic and real images.

\renewcommand{\arraystretch}{1.1} 

\begin{table}[h]
  \begin{center}
      \begin{tabular}{@{}@{}@{}@{}@{}@{}cccc@{}@{}@{}@{}@{}@{}}
	\toprule
	\multirow{2}{*}{Dataset} & \multicolumn{2}{c}{T-Less (primesense)} & 
	\multirow{2}{*}{SyntheT-Less} \\
	\cline{2-3}
	& Train & Test & ~\\
	\hline
	Number of samples & 38K & 10K & 30K \\
	Illumination variation & None & Small & Strong \\
	Occlusion & No & Yes & Yes \\ 
	Multi-objects images & No & Yes & Yes \\ 
	Object color variation & None & Small & Small \\ 
	Background variation & None & Small & Strong\\
        \bottomrule
      \end{tabular}
  \end{center}
  \vspace{-0.6cm}
    \caption{Comparison between the T-Less dataset and our SyntheT-Less dataset. }
    \label{tab:compare_datasets}
\end{table}

\subsection{Effectiveness of our Approach}

\begin{figure}[t]
  \begin{center}
      \begin{tabular}{cccc}
	\icg{0.096}{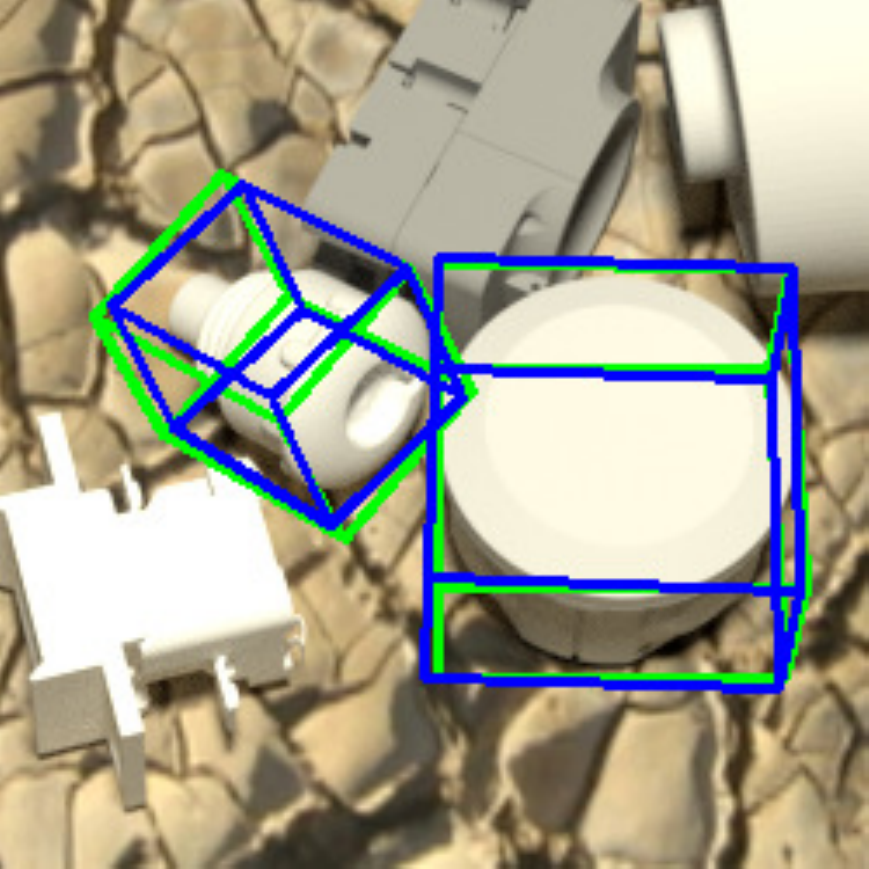} & 
	\icg{0.096}{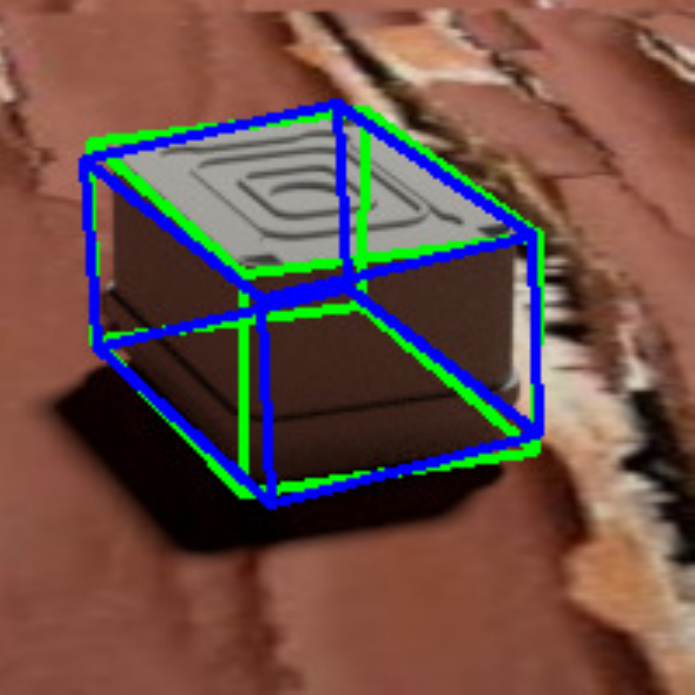} &
	\icg{0.096}{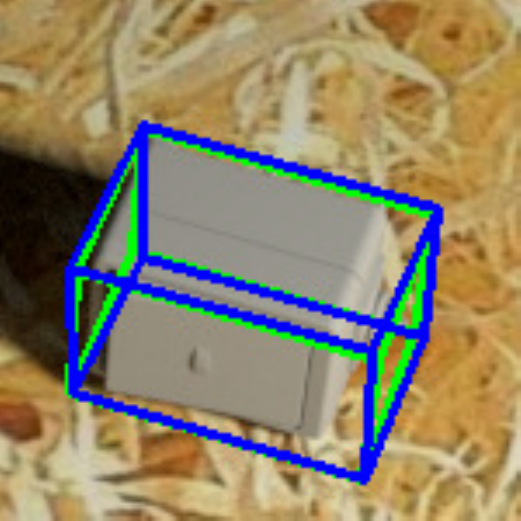} & 
	\icg{0.096}{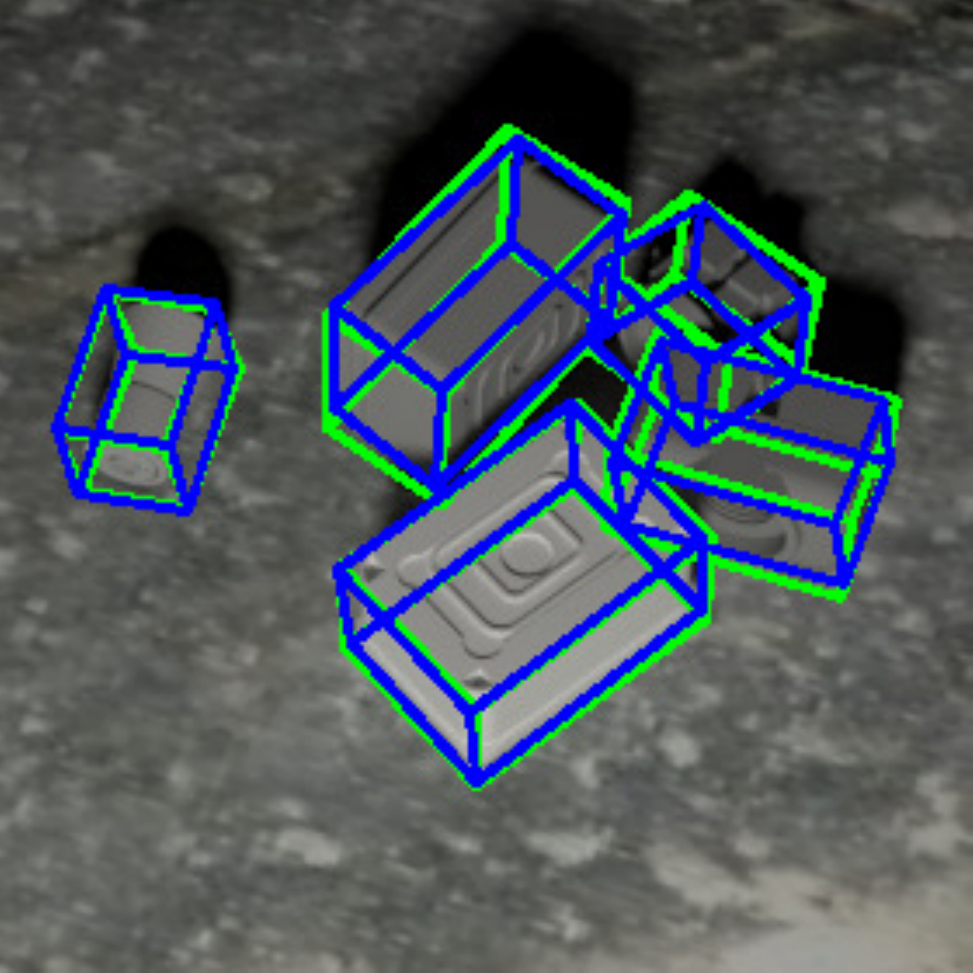} \\
	\icg{0.096}{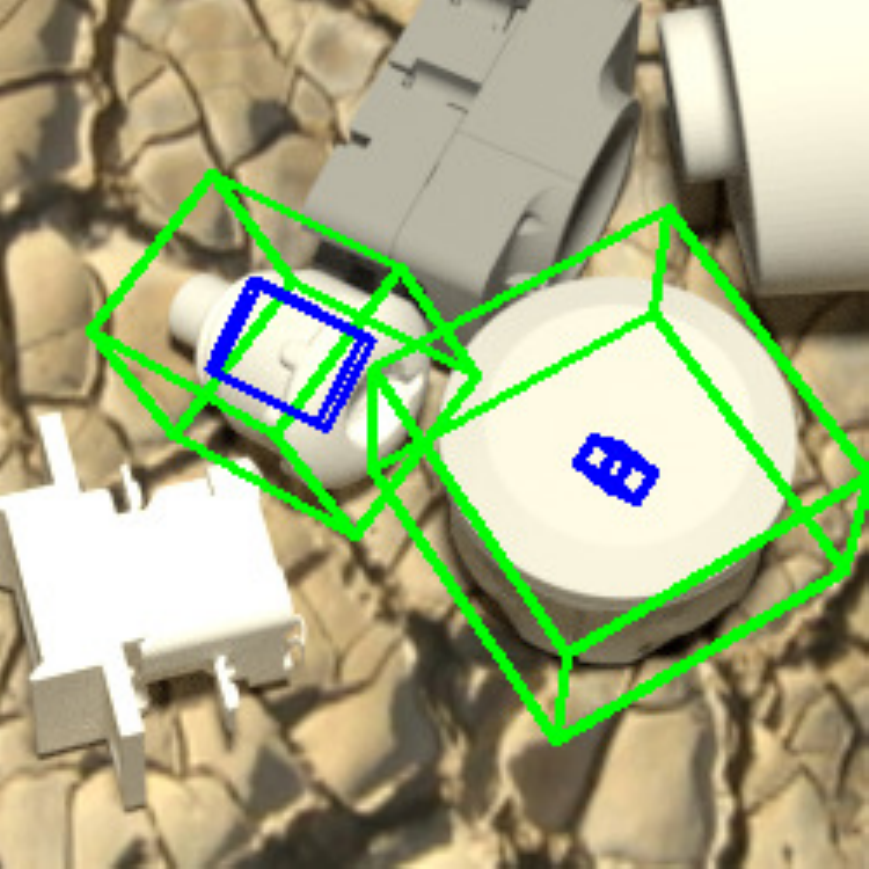} & 
	\icg{0.096}{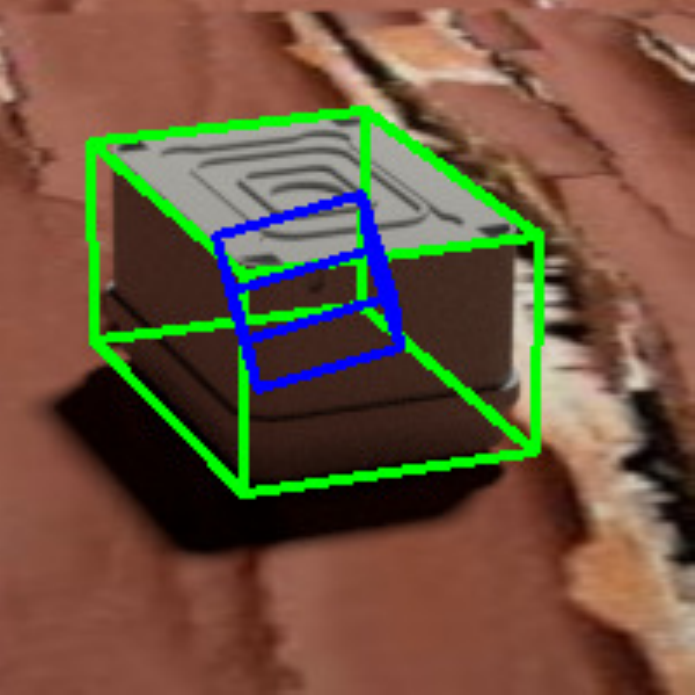} &
	\icg{0.096}{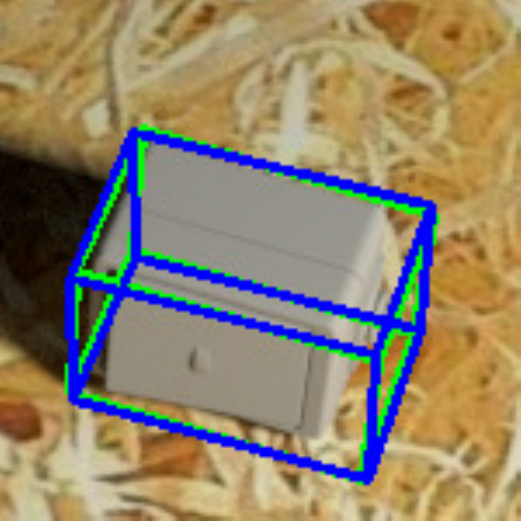} &
	\icg{0.096}{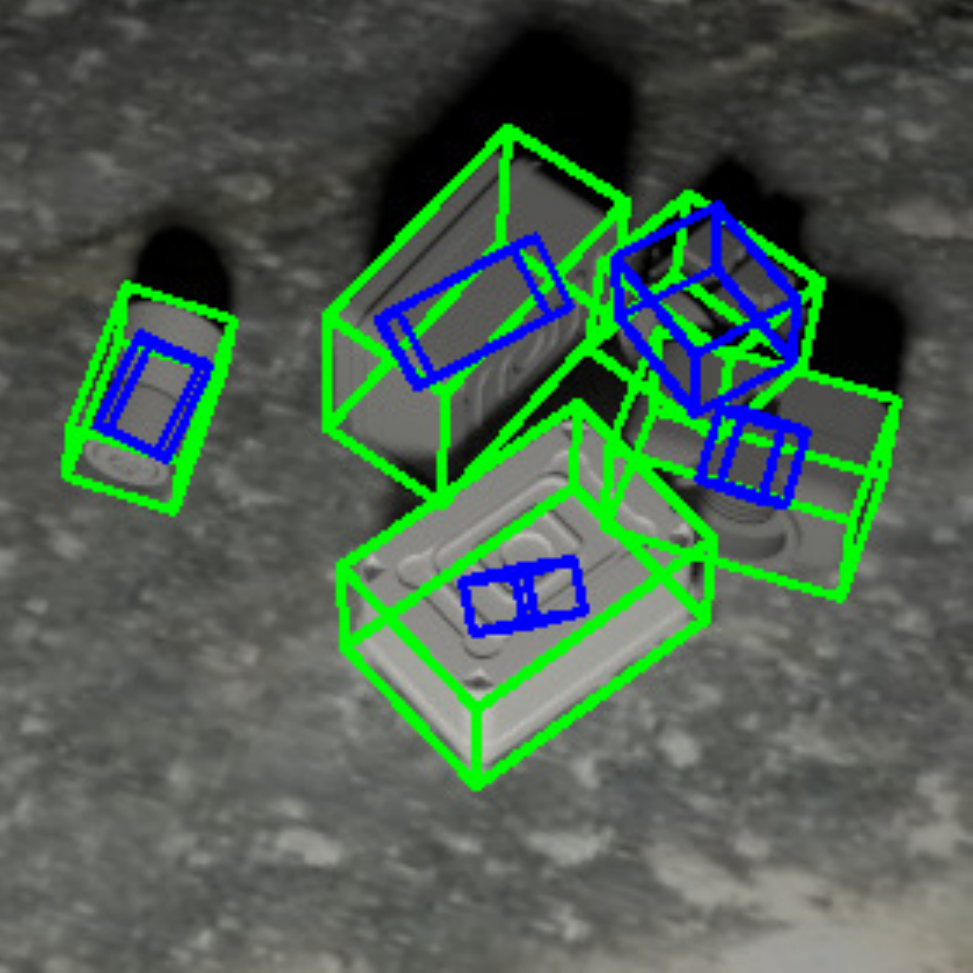}\\
	(a) & (b) & (c) & (d)
    \end{tabular}
    \vspace{-0.6cm}
    \caption{Pose estimation results with (top row) and without (bottom row) our
      normalization approach for  (a) generalized cylinders, (b)  an object with
      an axis of symmetry, (c) an object without any symmetry, and (d) a typical
      scene from our  SyntheT-Less dataset. The green and blue bounding boxes 
      correspond to the  ground truth poses estimated poses respectively.   
      Without our  normalization,  the network  learns  to predict  the
      average between all  the possible poses for symmetrical  objects, which is
      of course meaningless.}
    \label{fig:normalization_effectiveness}
  \end{center}
\end{figure}

\begin{figure}
\begin{center}
    \includegraphics[width=\linewidth, height= 
    7em]{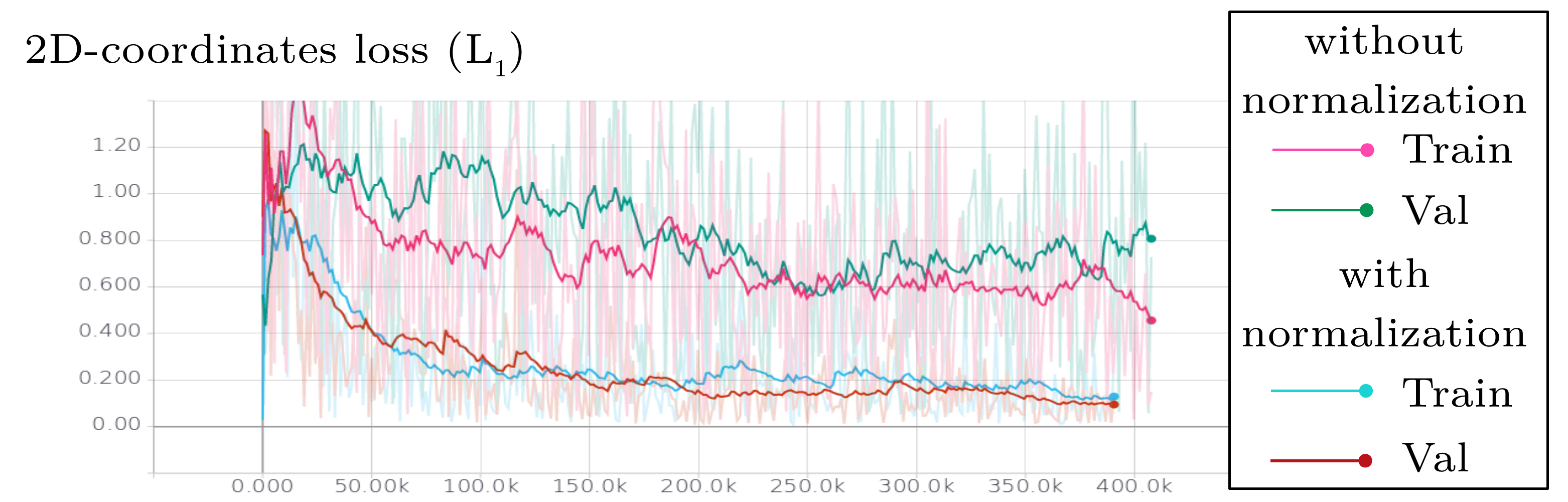}
    \vspace{-0.9cm}
  \caption{
    Learning curves  on the training  and validation  sets of our  Faster-RCNN 
    based
    implementation.   Without our  normalization described in 
    Section~\ref{sec:Method}, the network  fails to
    converge  to a  satisfying  solution.  More  exactly, it  converges  to a  local
    minimum  where  all  keypoints  collapse  at  the  center  of  the  object---see
    Fig.~\ref{fig:normalization_effectiveness}.  }
  \label{fig:loss_compare_normalisation}
 \end{center}
\end{figure}

As shown  in Fig.~\ref{fig:loss_compare_normalisation},  the loss of  our Faster
R-CNN -based  implementation converges  only when  the rotations  are normalized
using our normalization procedure, indicating that something is incorrect in the
loss function in absence of normalization.
In Fig.~\ref{fig:normalization_effectiveness}, we show  what happens in practice
for three possible types of objects: Two generalized cylinders (objects $30$ and
$3$), an object  with an axis of  symmetry (object $29$), and  an object without
any  symmetry (object  $26$).  When  dealing with  non-symmetrical objects,  the
network  is  able to  learn  the  6D pose  with  and  without the  normalization
procedure.   On the  opposite, when  the  objects are  symmetrical, without  our
normalization  the network  learns the  average between  all the  possible poses
ending up predicting a pose collapsed to the center of the object.

\begin{figure*}[t]
  \begin{center}
      \begin{tabular}{cccc}
	\icg{0.225}{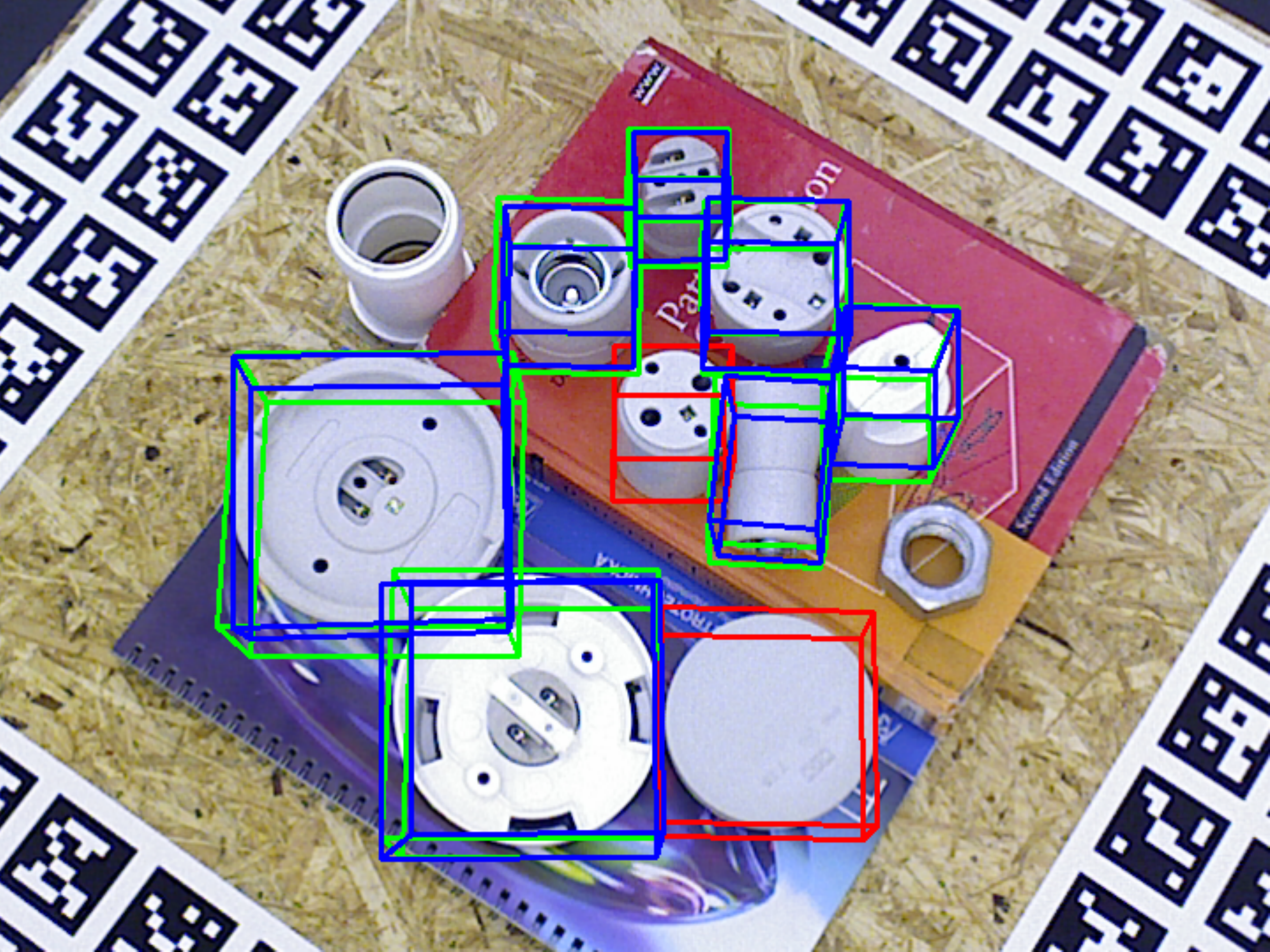} & 
	\icg{0.225}{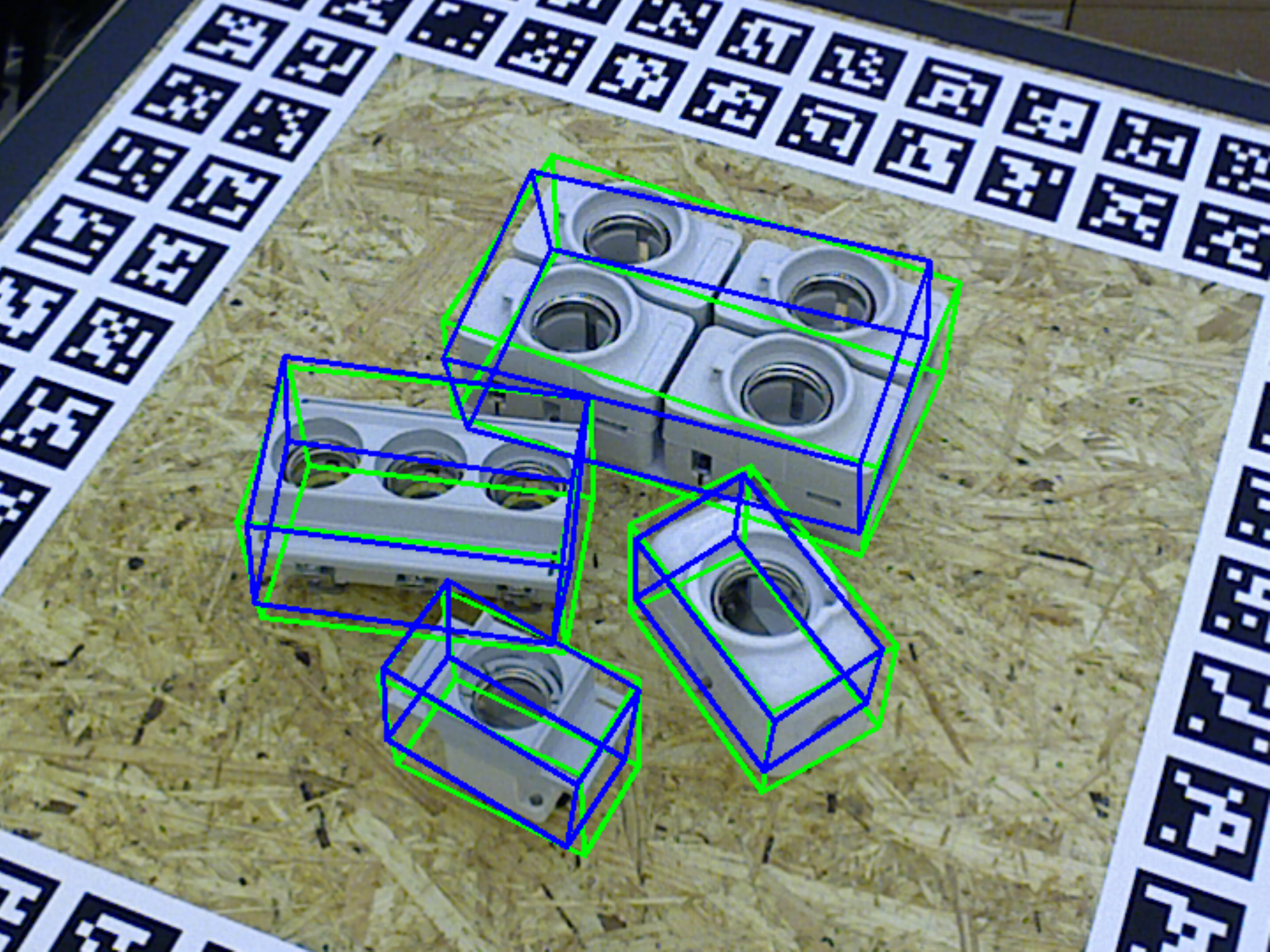} &
	\icg{0.225}{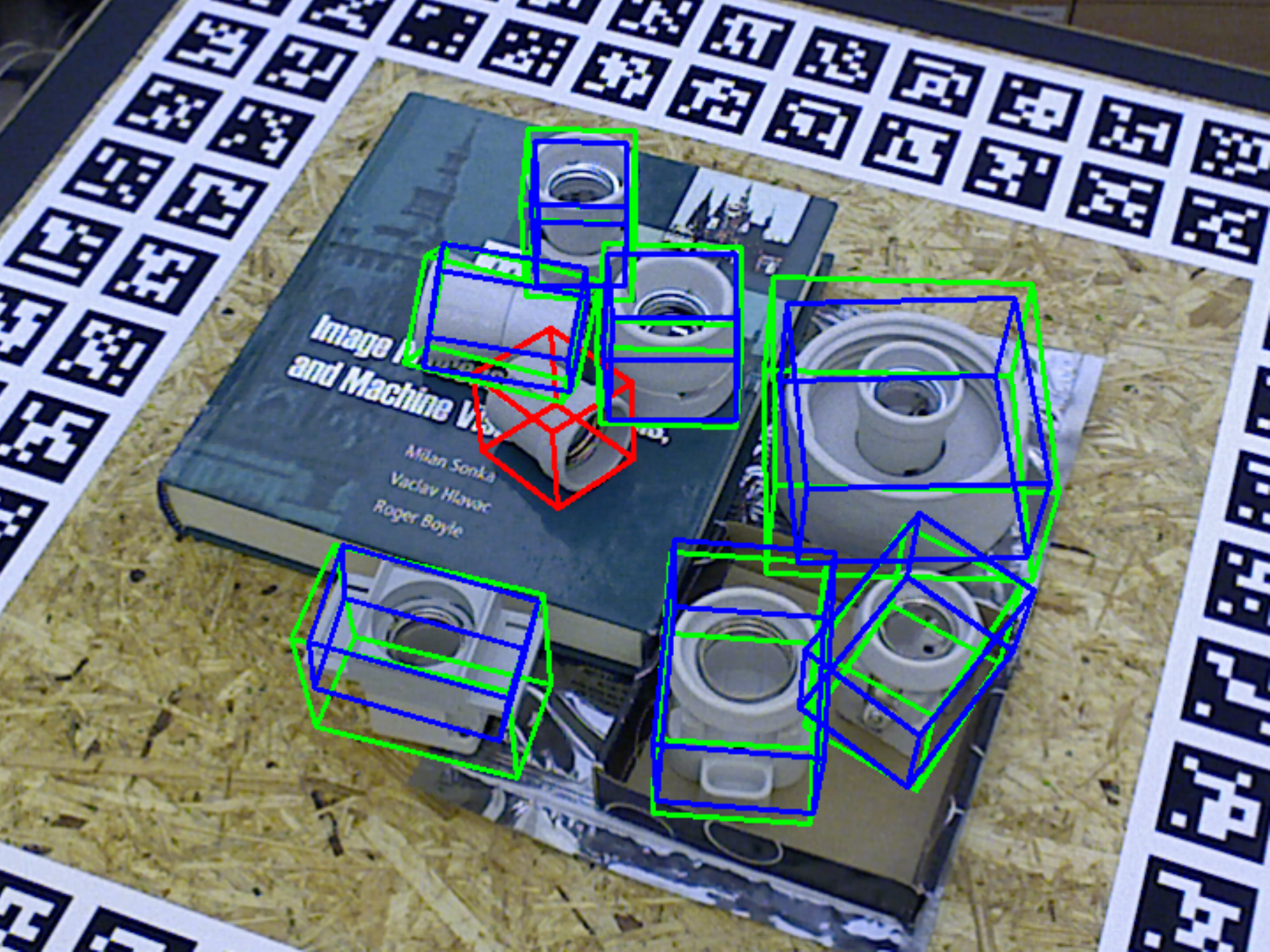} & 
	\icg{0.225}{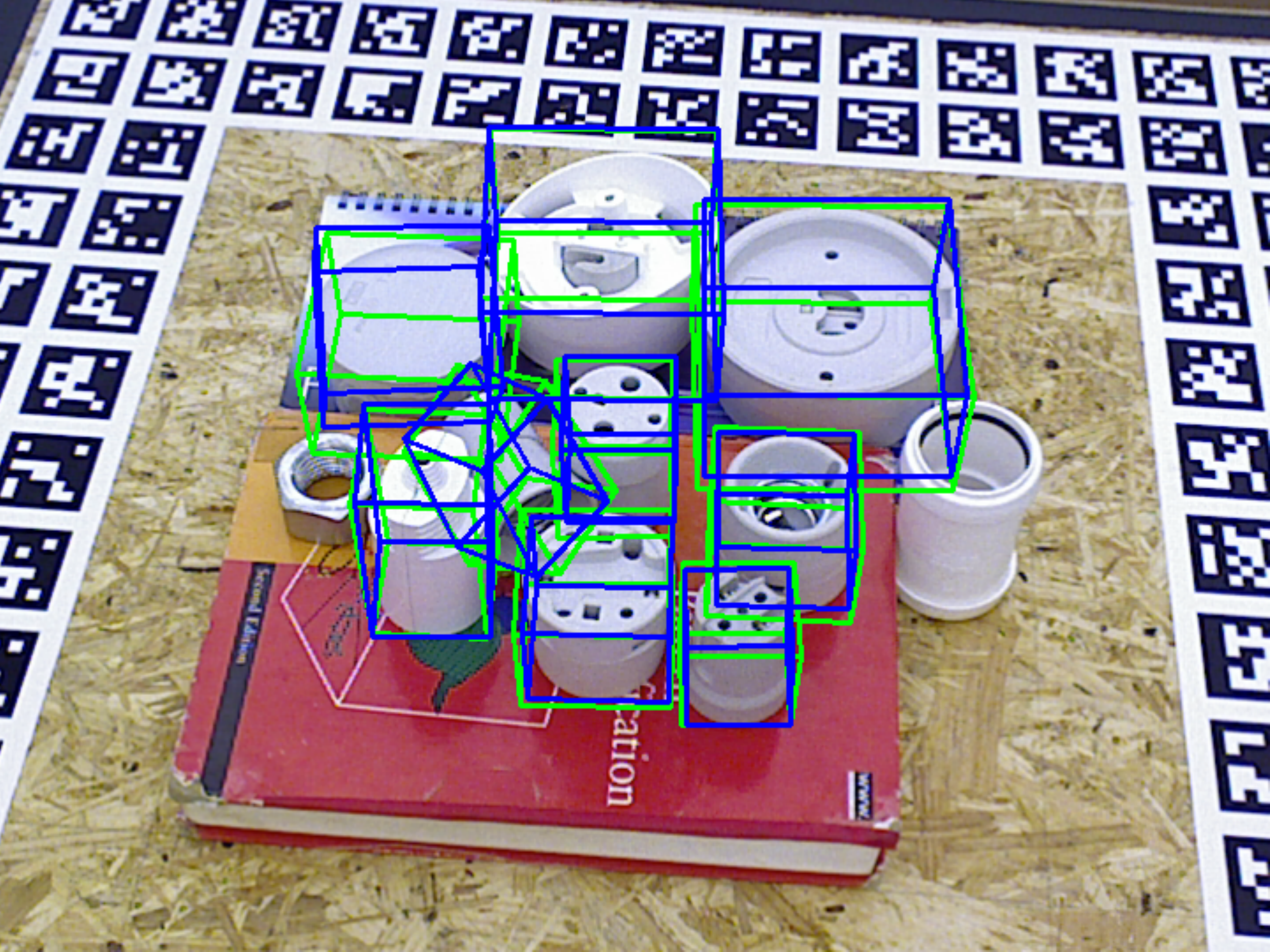} \\			 
	\icg{0.225}{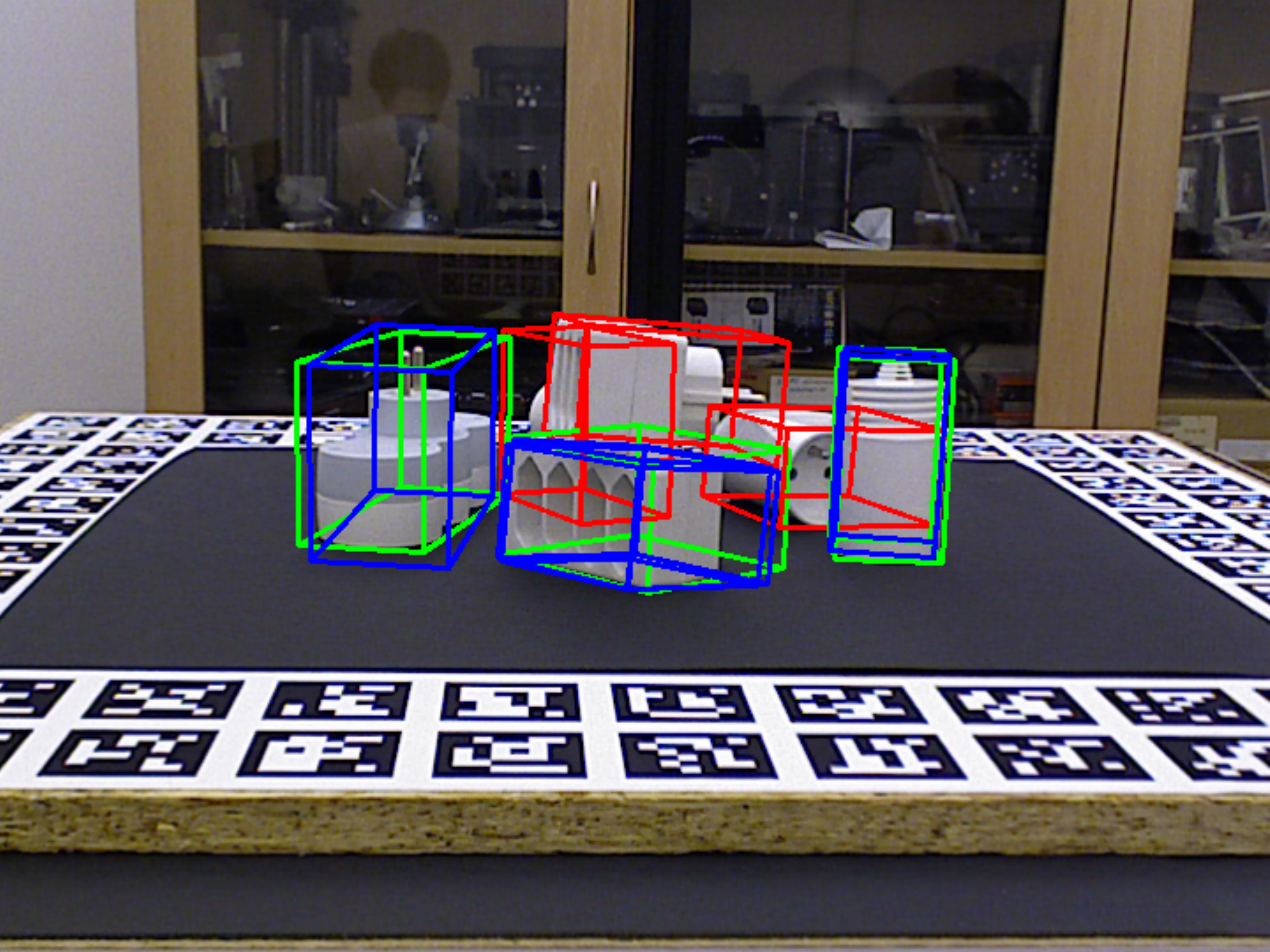} &
	\icg{0.225}{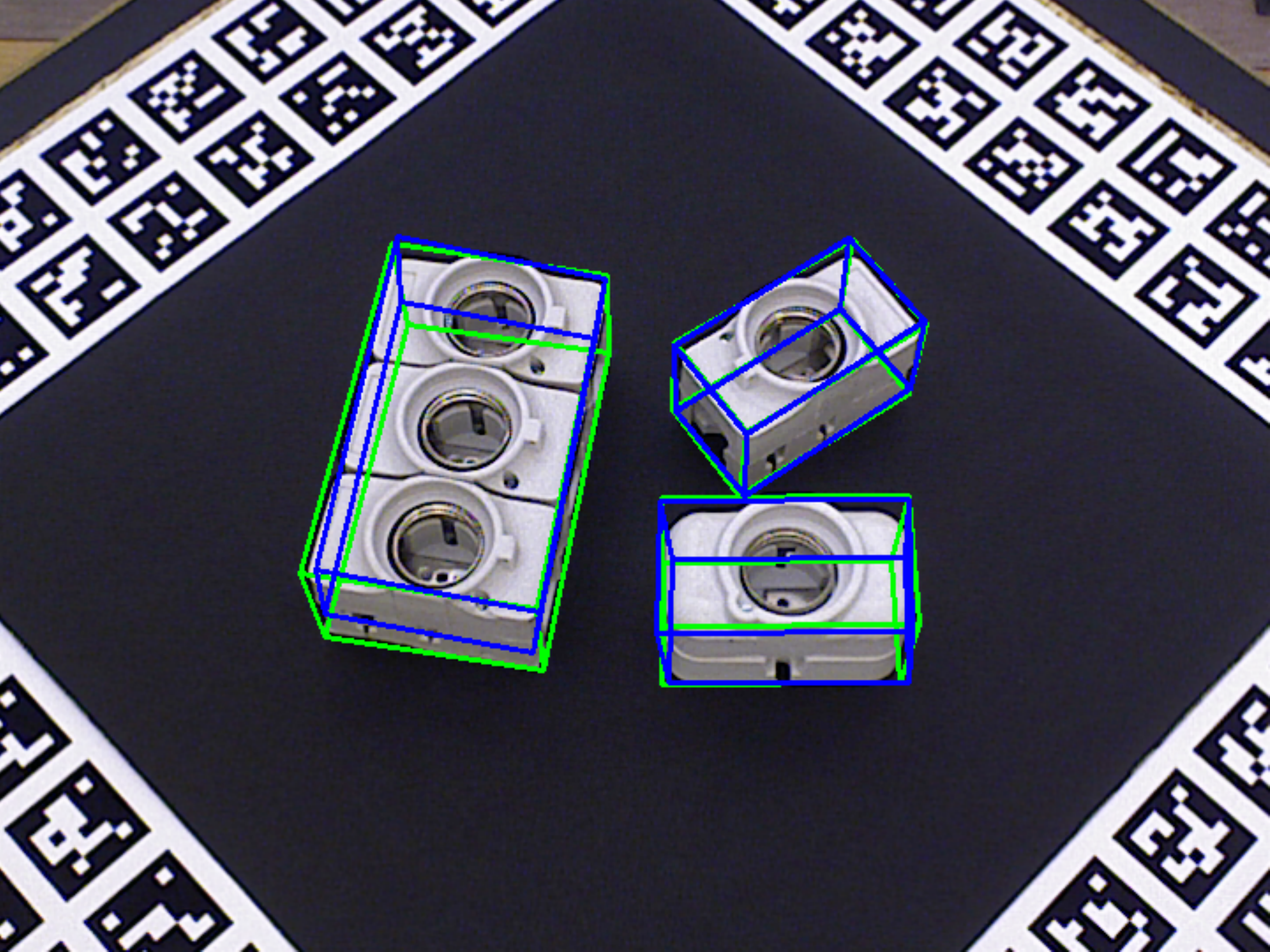} &
	\icg{0.225}{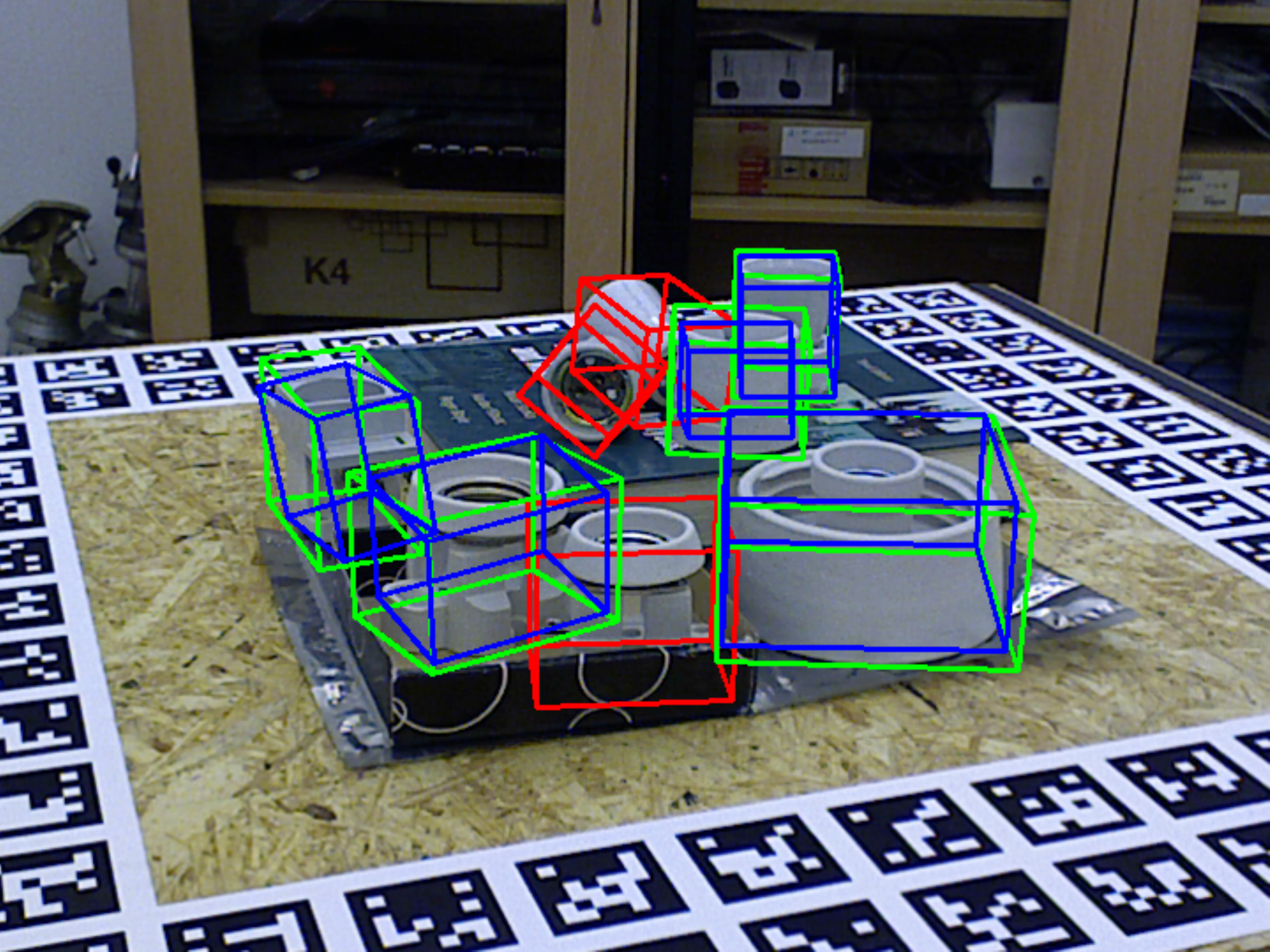} &
	\icg{0.225}{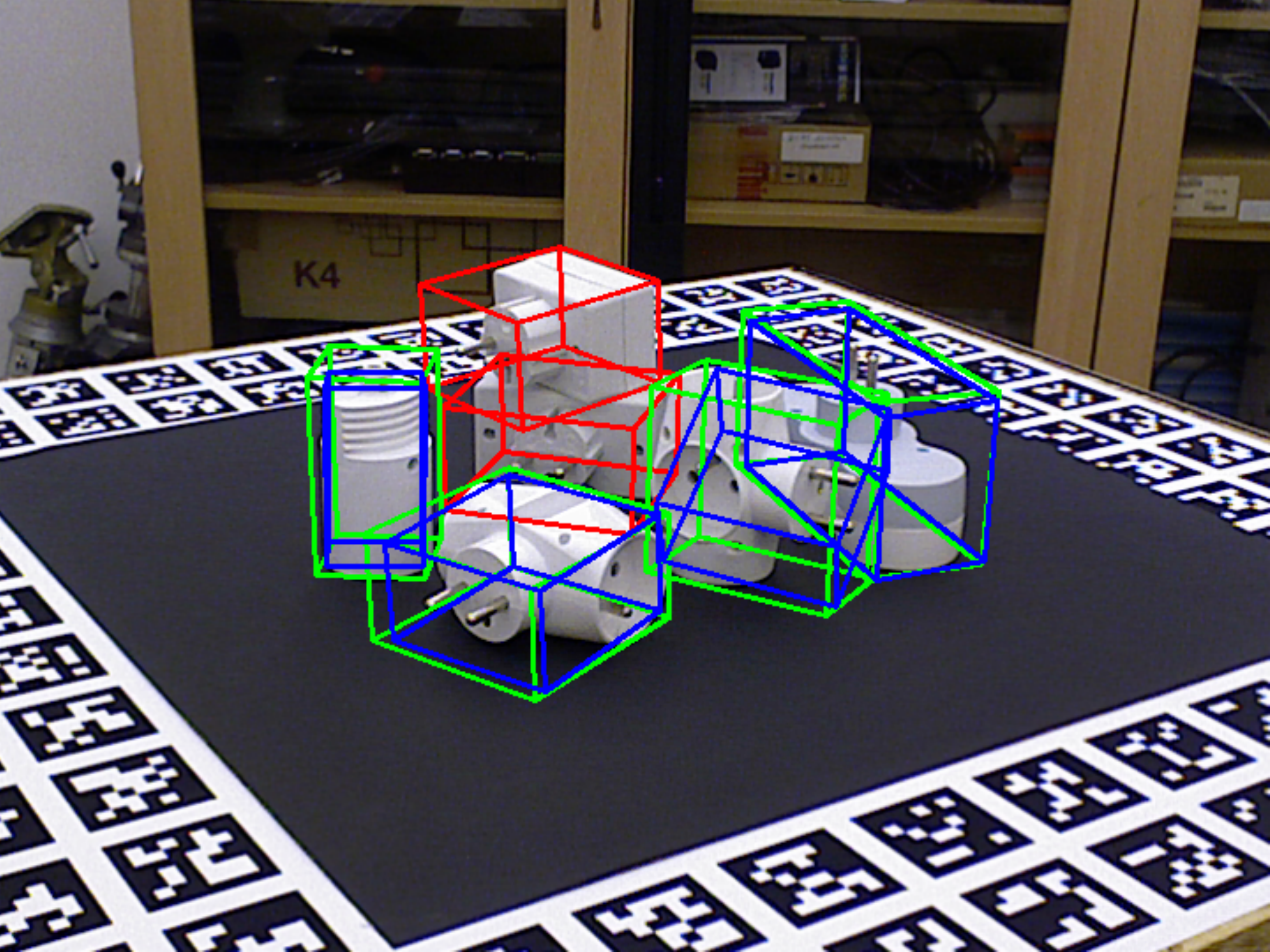}
    \end{tabular}
    \vspace{-0.6cm}
    \caption{Some qualitative results on test scenes of the T-Less
      dataset. Green and blue bounding boxes are the ground truth
      and estimated poses respectively while the red bounding boxes
      correspond to missed detections. }
    \label{fig:qualitative_results}
  \end{center}
\end{figure*}

\subsection{T-LESS Dataset: Comparison with \cite{Sundermeyer18}}

We use  the \textit{Visible  Surface Discrepancy} (VSD) error function  
introduced by \cite{Hodan18}. It compares  the ground truth measured depth  
maps $\hat{S}$ and the depth maps  $\bar{S}$ rendered according to the 
estimated  poses to evaluate
the proportion of visible pixels for which the depth absolute discrepancy map
\mbox{$|\hat{S}$   -  $\bar{S}|$}   is  below   a  threshold   $\tau$.   As   in
\cite{Hodan18},  we set  $\tau=20$mm and report  the
recall of correct 6D object poses at  $\mathbf{e_{vsd} < 0.3} $.  This metric is
not sensitive to  visual symmetries, as they induce similar  symmetries in depth
maps.

Table~\ref{tab:tless} compares our method to  the method of Sundermeyer \emph{et
  al}~\cite{Sundermeyer18}.  The object 3D orientation and translation along the
$\bx$-and  $\by$-axes  are  typically  well estimated.   Although  most  of  the
translation error is along $\bz$-axis, it is unsurprising since we do not use or
regress the depth information.  In order  to have a meaningful evaluation of our
results in terms of VSD, we keep the ground truth
of the translation along $\bz$-axis in our pose predictions.


\renewcommand{\arraystretch}{1.1}

\begin{table}
  \scriptsize
  \centering
  \begin{tabular}{@{}rccccc@{}}
    \toprule
    ~ & \multicolumn{3}{c}{Sundermeyer \textit{et
        al.}~\cite{Sundermeyer18}} && Ours \\
    \cmidrule{2-4}\cmidrule{6-6} Object & SSD & Retina & GT BBox && Faster-RCNN \\
    \hline 
    1 &  5.65 & 8.87 & {12.33}  && 26.35 \\ 
    2 &  5.46 &  { 13.22}  & 11.23  && 56.14 \\ 
    3 &  7.05 & 12.47 & {13.11}  && 83.33 \\ 
    4 &  4.61 & 6.56 & {12.71}  && 32.98 \\ 
    5 &  36.45 & 34.80 & {66.70}  && 44.54 \\ 
    6 &  23.15 & 20.24 & {52.30}  && 98.33 \\ 
    7 &  15.97 & 16.21 & {36.58}  && 87.74 \\ 
    8 &  10.86 & 19.74 & {22.05}  && 17.09 \\ 
    9 &  19.59 & 36.21 & {46.49}  && 52.54 \\ 
    10 &  10.47 & 11.55 & {14.31}  && 5.43 \\ 
    11 &  4.35 & 6.31 & {15.01}  && 27.97 \\ 
    12 &  7.80 & 8.15 & {31.34}  && 43.08 \\ 
    13 &  3.30 & 4.91 & {13.60}  && 48.54 \\ 
    14 &  2.85 & 4.61 & {45.32}  && 42.19 \\ 
    15 &  7.90 & 26.71 & {50.00}  && 47.10 \\ 
    16 &  13.06 & 21.73 & {36.09}  && 42.18 \\ 
    17 &  41.70 & 64.84 & {81.11}  && 56.83 \\ 
    18 &  47.17 & 14.30 & {52.62}  && 19.31 \\ 
    19 &  15.95 & 22.46 & {50.75}  && 27.53 \\ 
    20 &  2.17 & 5.27 & {37.75}  && 32.16 \\ 
    21 &  19.77 & 17.93 & {50.89}  && 41.19 \\ 
    22 &  11.01 & 18.63 & {47.60}  && 49.10 \\ 
    23 &  7.98 & 18.63 & {35.18}  && 26.08 \\ 
    24 &  4.74 & 4.23 & {11.24}  && 41.34 \\ 
    25 &  21.91 & 18.76 & {37.12}  && 44.37 \\ 
    26 &  10.04 & 12.62 & {28.33}  && 23.80 \\ 
    27 &  7.42 & 21.13 & {21.86}  && 33.78 \\ 
    28 &  21.78 & 23.07 & {42.58}  && 35.10 \\ 
    29 &  15.33 & 26.65 & {57.01}  && 15.92 \\ 
    30 &  34.63 & 29.58 & {70.42}  && 36.17 \\ 
    \hline Mean &  14.67 & 18.35 & {36.79}  &&  41.27 \\\bottomrule
  \end{tabular}
  \caption{T-LESS: Object recall for $err_{vsd}<0.3$ on all Primesense test 
    scenes (the higher the better).}
  \label{tab:tless}
\end{table}

\section{Conclusion}

In this paper, we studied the subtle problems that arise when training a machine
learning method to predict  the 6D pose of an object  with symmetries. This 
leads to a  simple method that  is agnostic to the  exact pose representation  
and the pose prediction  model.  Our  method can  therefore be  included in  
current and future developments for properly handling objects with symmetries. 
A direct extension of our work could be to automatically detect the object 
symmetries.

{\small
	\bibliographystyle{ieee}
	\bibliography{string,biblio}
}

\end{document}